\documentclass[journal]{IEEEtran}
\usepackage{breqn}

\usepackage{cite}
\usepackage{multirow}
\usepackage{array}
\usepackage{graphicx}
\usepackage{subcaption}

\usepackage{makecell} 
\usepackage{booktabs}
\usepackage{url}
\usepackage{tikz}
\usepackage{amssymb}
\usepackage{amsmath}
\usepackage[colorlinks=true,linkcolor=blue,urlcolor=blue,citecolor=blue]{hyperref}
\usepackage[table]{xcolor}

\begin{document}
%


\title{SAP-CoPE: Social-Aware Planning using Cooperative Pose Estimation with Infrastructure Sensor Nodes}

%
%
%

\author{ Minghao~Ning$^{1}$\textsuperscript{\textdagger},
         Yufeng~Yang$^{1}$\textsuperscript{\textdagger*},
         Shucheng~Huang$^{1}$,
         Jiaming~Zhong$^{1}$,
         Keqi~Shu$^{1}$,
         Chen~Sun$^{2}$,
         Ehsan~Hashemi$^{3}$,~\IEEEmembership{Senior Member,~IEEE},~and
         Amir~Khajepour$^{1}$,~\IEEEmembership{Senior Member,~IEEE}
\thanks{\textsuperscript{\textdagger} indicates equal contribution}
\thanks{* indicates the corresponding author}
\thanks{$^{1}$Minghao Ning, Yufeng Yang, Shucheng Huang, Jiaming Zhong, Keqi Shu and Amir Khajepour are with the Mechanical and Mechatronics Eng. Department, University of Waterloo, 200 University Ave W, Waterloo, ON N2L3G1, Canada. e-mail:\{minghao.ning, f248yang, s95huang, j52zhong, keqi.shu, a.khajepour\}@uwaterloo.ca}
\thanks{$^{2}$C. Sun is with the Department of Data and Systems Engineering, University of Hong Kong, Pok Fu Lam, Hong Kong, China (e-mail:c87sun@hku.hk)}
\thanks{$^{3}$Ehsan Hashemi is with the Mechanical Engineering Department, University of
Alberta, Alberta, T6G1H9, Canada (e-mail:ehashemi@ualberta.ca)}}

\newcommand\copyrighttext{%
  \footnotesize \textcopyright 2026 IEEE. Personal use of this material is permitted.
  Permission from IEEE must be obtained for all other uses, in any current or future
  media, including reprinting/republishing this material for advertising or promotional
  purposes, creating new collective works, for resale or redistribution to servers or
  lists, or reuse of any copyrighted component of this work in other works.
  }
\newcommand\copyrightnotice{%
\begin{tikzpicture}[remember picture,overlay]
\node[anchor=south,yshift=5pt] at (current page.south) {\fbox{\parbox{\dimexpr\textwidth-\fboxsep-\fboxrule\relax}{\copyrighttext}}};
\end{tikzpicture}%
}

\maketitle
\copyrightnotice

\begin{abstract}

Autonomous mobile robots must operate smoothly in human-populated indoor environments, where onboard-only sensing suffers from limited perception range and occlusions, and socially compliant motion planning must account for human psychological comfort zones. These factors complicate accurate recognition of human intentions and the generation of comfortable, social-aware trajectories.
To address these challenges, we propose SAP-CoPE, an indoor navigation framework that integrates cooperative infrastructure sensing, human pose estimation, and a social-aware MPC planner.
In the perception module, an uncertainty-aware optimization problem is formulated to estimate 3D human poses from camera observations, with optional incorporation of sparse LiDAR point-cloud measurements, while considering human joint coherence. In the planning module, the estimated human pose and orientation are used to construct a personal-space (PS) field within an MPC framework, enabling the robot to account for both physical safety and psychological comfort.
Extensive real-world evaluations demonstrate the effectiveness of the proposed approach in generating social-aware trajectories for autonomous systems. 
\end{abstract}

\def\abstractname{Note to Practitioners}
\begin{abstract}
Indoor autonomous mobile robots operating in public facilities such as hospitals must navigate crowded environments where onboard sensing is often challenged by occlusions and limited fields of view, while also accounting for human comfort in addition to collision avoidance. This paper presents SAP-CoPE, a deployable framework that integrates cooperative infrastructure sensing with social-aware motion planning. Ceiling-mounted sensor nodes provide expanded and occlusion-robust perception through uncertainty-aware 3D human pose estimation from cameras, optionally supplemented by sparse LiDAR data. Leveraging these pose estimates, an MPC-based planner incorporating a human personal space field generates trajectories that balance social comfort, efficiency, and safety while respecting the robot’s kinematic constraints. The proposed architecture is modular and compatible with existing indoor robotic platforms using standard hardware, and real-world experiments demonstrate smoother navigation and improved clearance and travel-time performance compared with conventional baseline methods.
\end{abstract}

\begin{IEEEkeywords}
Cooperative Perception, Human Pose Estimation, Social-awareness, Motion Planning, Model Predictive Control (MPC).
\end{IEEEkeywords}


%
\IEEEpeerreviewmaketitle

\section{Introduction}

\begin{figure}[t] \centering \includegraphics[width=0.45\textwidth]{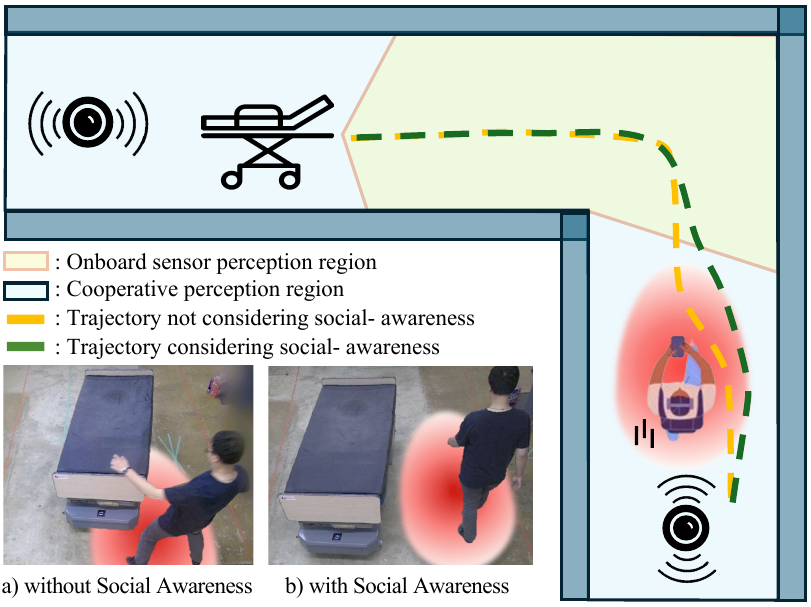} \vspace*{-0.1cm} \caption{Key challenges: (1) How to reliably detect humans in crowded, occluded scenarios? (2) How to guide the robot for human-comfortable motion?} \label{fig:problem_to_solve} \vspace*{-0.4cm} \end{figure}

\IEEEPARstart{I}ntelligent indoor mobility systems are gaining increasing attention in healthcare, logistics, and the automotive industry \cite{law2021case, pikner2021cyber, savci2022improving}. These systems reduce manual transportation tasks, minimize workplace injuries, and improve operational efficiency \cite{kim2021analysis}. In restricted environments, mobile robots primarily focus on avoiding collisions with objects and are programmed to stop when humans are nearby. However, this strategy is less effective in public areas like hospitals and shopping malls, where interactions with people are unavoidable. As shown in Fig.~\ref{fig:problem_to_solve}, to enable safe and efficient human-robot interaction, such a system requires real-time obstacle detection, human state and motion-cue estimation, and a robust motion planning algorithm for safe and smooth navigation.

Recent research has improved robot perception using high-performance onboard sensors; however, onboard-only systems remain limited by perception range, field of view (FOV), and occlusions in crowded indoor spaces \cite{schmid2023dynablox}. Infrastructure-supported perception can mitigate these limitations by observing shared spaces from complementary viewpoints. However, for human-aware navigation, object-level information such as pedestrian positions or bounding boxes is often insufficient, because it provides limited cues about human body orientation, posture, and short-term motion tendency.

Articulated 3D human pose provides richer information for social-aware planning. Human heading and body configuration can indicate whether a pedestrian is facing the robot, crossing its path, or beginning to turn. Such cues are useful for constructing an orientation-aware personal-space (PS) field, rather than treating each human as an isotropic obstacle. This is particularly important for large indoor robotic platforms, whose rectangular footprints and kinematic constraints require the planner to balance human comfort, physical safety, travel efficiency, and motion feasibility.

This work builds upon our preliminary studies on infrastructure-supported perception and MPC-based indoor robot navigation \cite{ning2024enhancingindoormobilityconnected, yang2024intelligentmobilityintegratedmotion}. Those studies established the foundations for cooperative infrastructure perception and physically safe MPC navigation, but they focused mainly on obstacle-level perception and collision-oriented planning. In contrast, this article develops a unified pose-aware and social-aware navigation framework. The proposed SAP-CoPE framework estimates human 3D pose and heading in the world coordinate system using uncertainty-aware camera projection, joint-coherence constraints, and optional sparse LiDAR associations. The estimated pose and heading are then incorporated into an MPC-based planner through an asymmetric PS field, enabling the robot to generate socially acceptable motion while satisfying its kinematic constraints.

The key contributions of this article are as follows.

\begin{enumerate}
\item \textit{Uncertainty-aware and sparse-measurement-compatible 3D human pose estimation:} A lightweight optimization-based pose estimation formulation is proposed to account for uncertainty propagation from camera projection while enforcing human joint coherence. The sparse LiDAR term is activated only when reliable LiDAR-to-joint associations are available, allowing the method to operate in both camera-only and camera-LiDAR configurations.

\item \textit{Extension from physically safe MPC to social-aware MPC navigation:} A human-orientation-dependent PS field is incorporated into the MPC formulation for a large omnidirectional indoor robot. Instead of treating humans only as generic obstacles, the planner uses estimated human pose and heading to generate socially acceptable motion that balances clearance, comfort, travel efficiency, and robot kinematic feasibility.

\item \textit{Practical validation and scope demonstration in human-populated indoor environments:} The proposed SAP-CoPE framework is validated through simulation and real-world experiments with human participants. The results demonstrate its applicability to infrastructure-supported indoor autonomous mobility tasks, where the robot must navigate efficiently while maintaining human-comfortable interaction distances.

\end{enumerate}

The remainder of this article is organized as follows. Section~II reviews related work. Section~III presents an overview of the proposed methodology. Section~IV details the perception algorithm. Section~V formulates the social-aware motion planner. Section~VI presents the experimental results. Finally, Section~VII concludes the paper and discusses future research directions.

\section{Related Work}


\subsection{Perception} 

\textbf{Infrastructure-Based Perception Systems:}
Reliable perception is fundamental for intelligent indoor mobility systems. Traditional infrastructure-based approaches, particularly those using vision sensors, have been widely explored. Works such as \cite{zhou2012understanding,heya2018image,haque2018visionbased,carro2023multicamera} employ ceiling or wall-mounted cameras to detect and track pedestrians. However, purely independent camera-driven systems can be sensitive to lighting conditions and occlusions, and they are generally limited in their ability to obtain precise 3D object positions without additional depth information.
To overcome these limitations, researchers have explored multi-sensor fusion within the infrastructure. Br{\v{s}}{\v{c}}i{\'c} et al. \cite{brvsvcic2013person} integrated RGB-D cameras, LiDAR, and marker-based tracking to enhance spatial accuracy and robustness. Though this approach improves performance, its scalability is hindered by high equipment and installation costs. Similarly, high-frequency motion capture systems \cite{rudenko2020thor} offer ground-truth level accuracy but are restricted to controlled spaces, limiting their practical use in real-world deployments. 
Overall, these limitations emphasize the demand for indoor perception systems that are both reliable under challenging conditions and feasible for cost-effective, large-scale deployment.

\textbf{Cooperative Perception Systems:}
Cooperative perception improves environmental awareness by integrating sensory data from multiple sources, including other robots and infrastructure. This collaborative approach effectively addresses challenges such as occlusions and the limited FOV inherent in individual sensors. Recent advancements highlight the significance of Vehicle-to-Everything (V2X) in the autonomous driving domain\cite{ADNANYUSUF2024100980, huang2023v2x, bai2024survey, caillot2022cooperative}. 
Huang et al. \cite{huang2023v2x} highlight the critical role of data sharing between agents and infrastructure while identifying real-time performance and communication delay as key bottlenecks.
Inspired by this concept, we propose a real-time cooperative perception system utilizing ceiling-mounted ISNs. Each node performs local perception and transmits lightweight observations to a central fusion module in real time. This architecture not only expands FOV and reduces occlusions in crowded indoor spaces but also ensures timely and coherent perception by mitigating latency-induced errors in the fusion pipeline.

\textbf{3D Human Pose Detection:}
Accurately estimating human 3D poses is vital for autonomous systems to interpret and predict human behaviour, ensuring safe and social-aware interactions. 
Single-camera-based methods face challenges such as depth ambiguity and occlusions \cite{wang2021deep}; therefore, estimation accuracy can be improved by leveraging multi-modal sensor data. 
Zanfir et al. introduced HUM3DIL, a semi-supervised multi-modal approach that embeds LiDAR points into pixel-aligned multi-modal features and employs a Transformer architecture for prediction\cite{zanfir2023hum3dil}. Similarly, Bauer et al. proposed a weakly supervised method that integrates camera and LiDAR data, enabling accurate 3D human pose estimation without extensive 3D annotations\cite{bauer2023weakly}. However, this method requires dense point clouds to generate high-accuracy pseudo-3D keypoint labels by projecting point clouds onto the image plane.
Different from learning-based RGB--LiDAR pose estimation methods such as HUM3DIL and Bauer et al., our method does not assume dense LiDAR observations or learned multimodal feature fusion. In our deployed infrastructure layout, a pedestrian is usually observed by one top-down camera, while LiDAR returns on the human body may be sparse or intermittently available. Therefore, we formulate 3D pose estimation as a lightweight probabilistic optimization problem in the world coordinate frame. The method explicitly propagates calibrated camera-projection uncertainty, enforces joint-coherence constraints, and incorporates LiDAR only as optional weighted geometric residuals when valid associations exist. This allows the same framework to operate in camera-only, camera--LiDAR, single-node, and cooperative configurations.

\subsection{Planning}
\textbf{social-aware Planning:}
Recent human-aware planning algorithms focus primarily on modeling human–robot interactions while often assuming the robot has a simple circular shape for computational convenience, without explicitly accounting for rectangular or articulated platform geometries \cite{chen2023trajectory,fang2023unified, shu2023human, shu2024game}. Among these algorithms, the Social Force Model (SFM) is widely used to generate socially acceptable robot motion. Kamezaki et al. \cite{kamezaki2022reactive} introduced a planning algorithm based on the inducible Social Force Model, while Ratsamee et al. \cite{ratsamee2013human} developed the Extended SFM (ESFM) by incorporating human pose and orientation to improve conventional SFM. Despite their effectiveness, SFM-based approaches often model human influence as symmetric in all directions. While this assumption allows the robot to maintain distance from personal spaces in sparse environments, it can lead to frequent unexpected deceleration or even stopping in crowded settings, thereby reducing efficiency and increasing the risk of collisions.

\textbf{Planning based on Information Types:}
 Social-aware planning algorithms can be broadly categorized into two types: individual state-based information and group-based information \cite{truong2017toward}. Individual state-based methods use detailed information such as position, pose, orientation, and velocity; for example, Stefanini et al. \cite{stefanini2024efficient} proposed a novel planning algorithm that incorporates the articulated 3D human poses, semantic labels, and trajectory prediction to describe human behaviour. However, such approaches usually require reliable perception and tracking algorithms, and their performance can degrade significantly under occlusions, which are common in cluttered environments.
 
 In contrast, group-based methods treat multiple humans as a single entity. Hoang et al. \cite{ngo2022socially} proposed a proactive social motion model that accounts for socio-spatiotemporal characteristics of human groups, allowing the mobile robot to plan a safe trajectory in various situations. Nevertheless, group-level representations may oversimplify complex interactions among individuals, potentially causing the robot to react conservatively or ignore subtle cues such as interpersonal distances, leader–follower structures, or individuals breaking away from the group. 
 
On the other hand, although the above approaches have demonstrated strong performance in human-aware planning, they may not generalize well to large, rectangular robots with omnidirectional maneuverability, potentially resulting in unrealistic or impractical motions. Moreover, these methods often lack explicit constraints on robot kinematics and control limits, which may produce motion commands that are smooth in simulation but difficult to track reliably on actual hardware.

\section{Methodology}

 \begin{figure*}[t]
    \centering
    \includegraphics[width=0.95\textwidth]{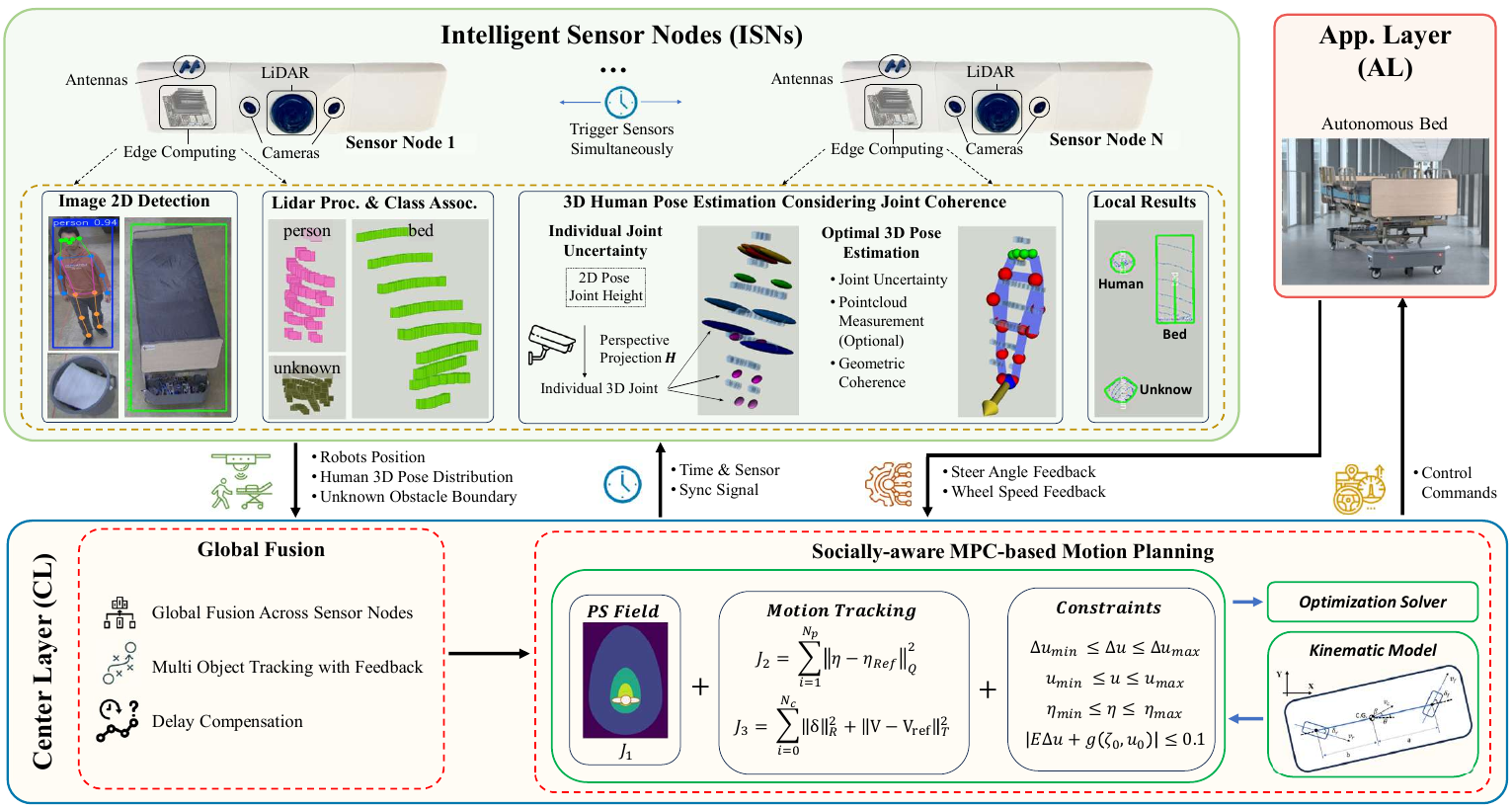}
    \vspace*{-0.2cm}
    \caption{SAP-CoPE Structure Overview}
    \label{fig:framework}
    \vspace*{-0.2cm}
\end{figure*}

The proposed SAP-CoPE method consists of two main components: real-time cooperative perception and social-aware motion planning (Fig. \ref{fig:framework}). The perception system employs a group of multi-modal ISNs equipped with LiDAR, cameras, and edge computing for local object detection and 3D human pose estimation. A synchronized clock system aligns LiDAR scans with camera operations, ensuring temporally consistent data, which is transmitted to a Central Layer (CL) for global fusion, multi-object tracking, and delay compensation. For the motion-planning system, an MPC-based planner is designed to generate optimal poses and is implemented in the CL, enabling the robot to compute a social-aware motion that respects humans' personal space while remaining close to a predefined reference motion. To achieve this, the PS field, inspired by psychological studies, is incorporated into the MPC’s objective function.

Furthermore, the controller then tracks the generated motion to obtain optimal control commands, which are transmitted to the Application Layer (AL) for execution. In this study, communication between the ISNs, CL, and AL is achieved using a Wi-Fi network. 



\section{Perception}
\subsection{Local Perception}
The local perception module consists of camera-based 2D detection, LiDAR point cloud processing, multi-modal class association, and a novel probabilistic 3D human pose estimation framework incorporating joint coherence constraints.
\subsubsection{Camera-based 2D Detection}
Recent advancements in deep learning have significantly improved camera-based 2D object detection. Notably, YOLOv11 \cite{yolov11_ultralytics} demonstrates superior accuracy and inference speed due to its state-of-the-art backbone and neck architectures. However, the pre-trained model does not generalize well to the top-down sensor node view and is unable to detect the autonomous robot used in our application. To address these limitations, we fine-tuned the YOLOv11 model using a dataset combining COCO \cite{lin2015microsoft} and the Indoor Cooperative Infrastructure Dataset (ICID) \cite{ning2024enhancingindoormobilityconnected} to achieve human 2D pose detection and robot bounding box detection in a single network.

\subsubsection{LiDAR Processing and Class Association}
The LiDAR processing consists of region-of-interest filtering and a hierarchical clustering algorithm to cluster obstacle points while accounting for the LiDAR scanning pattern. The clustering results are then fused with the camera-based 2D detection results for robot and human pose estimation.
More details on the implementation of clustering and association can be found in our previous work \cite{ning2024enhancingindoormobilityconnected}. The proposed 3D human pose estimation method is discussed in the following section.

\subsubsection{3D Human Pose Estimation with Joint Coherence Constraints}
Estimating absolute 3D human pose from a single camera is challenging due to the lack of depth information. As indicated in Fig. \ref{fig:framework}, our approach first obtains the 2D pixel coordinates of each joint and leverages prior height distribution information to estimate the rough 3D distribution of each joint as $P^i \sim \mathcal{N}(\mu_{P^i}, \Sigma_{P^i})$ in the absolute world coordinate frame. This estimation is performed using the camera projection matrix $H$, ensuring a more robust and coherent 3D pose representation. 

\paragraph{World Coordinate and Jacobian Calculation}
Given a pixel point \((x_p, y_p)\), height information \(z_w\), and the camera projection matrix \(H\), we aim to estimate the world coordinates \((x_w, y_w)\) and the Jacobian matrix \(\mathbf{J}\) of \((x_w, y_w, z_w)\) with respect to \((x_p, y_p, z_w)\). The camera projection formula is given by:
\begin{equation}
    s \begin{bmatrix} x_p \\ y_p \\ 1 \end{bmatrix} = K [R \ t] \begin{bmatrix} x_w \\ y_w \\ z_w \\ 1\end{bmatrix} = H_{3\times4} \begin{bmatrix} x_w \\ y_w \\ z_w \\ 1\end{bmatrix}
    \label{eq:camera-projection}
\end{equation}
where $s$ is a scale factor; $x_p$ and $y_p$ are the pixel coordinates, $K$ is the camera intrinsic matrix, $R$ is the rotation matrix, $t$ is the translation vector, $x_w$, $y_w$, $z_w$ are the world coordinates.

Given a projection matrix \(H \in \mathbb{R}^{3 \times 4}\), which maps the 3D world coordinates to the 2D image plane, we denote its elements as follows: 
\begin{equation}
H = \begin{bmatrix} h_{11} & h_{12} & h_{13} & h_{14} \\ h_{21} & h_{22} & h_{23} & h_{24} \\ h_{31} & h_{32} & h_{33} & h_{34} \end{bmatrix}
\end{equation}

We define the intermediate variables:
\begin{equation}
\begin{aligned}
a_{11} &= h_{31} x_p - h_{11}, \quad a_{12} = h_{32} x_p - h_{12}, \\
a_{21} &= h_{31} y_p - h_{21}, \quad a_{22} = h_{32} y_p - h_{22}, \\
b_1 &= h_{13} z_w - h_{33} x_p z_w - h_{34} x_p + h_{14}, \\
b_2 &= h_{23} z_w - h_{33} y_p z_w - h_{34} y_p + h_{24}.
\end{aligned}
\end{equation}

The world coordinates \((x_w, y_w)\) are then calculated as:
\begin{equation}
\begin{aligned}
x_w &= \frac{b_1 a_{22} - b_2 a_{12}}{a_{11} a_{22} - a_{12} a_{21}}=\frac{N_x}{D}, \\
y_w &= \frac{b_2 a_{11} - b_1 a_{21}}{a_{11} a_{22} - a_{12} a_{21}}=\frac{N_y}{D}.
\end{aligned}
\end{equation}

The Jacobian matrix \(\mathbf{J}\) of \((x_w, y_w, z_w)\) with respect to \((x_p, y_p, z_w)\) is given by:
\begin{equation}
\mathbf{J} = 
\begin{bmatrix}
\frac{\partial x_w}{\partial x_p} & \frac{\partial x_w}{\partial y_p} & \frac{\partial x_w}{\partial z_w} \\
\frac{\partial y_w}{\partial x_p} & \frac{\partial y_w}{\partial y_p} & \frac{\partial y_w}{\partial z_w} \\
0 & 0 & 1
\end{bmatrix}
\end{equation}

The complete derivation can be found in supplementary material.
This Jacobian matrix allows uncertainty to be propagated from the image plane to the world coordinate frame, which is essential for precise 3D pose estimation.



\paragraph{Propagation of Uncertainty to 3D World Coordinates}

Building upon the computed world coordinates and the derived Jacobian matrix, we
propagate the uncertainties of the input parameters $(x_{p}, y_{p}, z_{w})$ to estimate
the 3D distribution of each joint in the absolute world coordinate system.

Using the Jacobian matrix \(\mathbf{J}\), the covariance matrix of the estimated 3D position \((x_w, y_w, z_w)\) is obtained through uncertainty propagation as: $\Sigma_{P^i} = \mathbf{J} \cdot \Sigma_{(x_p, y_p, z_w)} \cdot \mathbf{J}^\top$, and the combined input uncertainty is defined as: $\Sigma_{(x_p, y_p, z_w)} = \mathrm{diag}(\sigma_{x_p}^2, \sigma_{y_p}^2, \sigma_{z_w}^2)$.


The resulting covariance matrix \(\Sigma_{P^i}\) denotes the uncertainty in the world coordinates of the joint \(P^i\). Combined with the mean position \(\mu_{P^i} = [x_w, y_w, z_w]^\top\), we represent the 3D position of each joint as a multivariate Gaussian distribution: $P^i \sim \mathcal{N}(\mu_{P^i}, \Sigma_{P^i})$.

\paragraph{Joint Coherence in 3D Pose Estimation}

To maintain coherence in the inferred 3D human pose, we impose additional constraints based on the spatial relationships between adjacent joints. This is achieved by incorporating prior knowledge of bone lengths and their associated uncertainties. For example, given two adjacent joints \(P^i\) and \(P^j\), the constraint on their Euclidean distance is modeled as $\| \mu_{P^i} - \mu_{P^j} \| \approx l_{ij}$,
where \(l_{ij}\) is the prior mean of the bone length between the joints \(i\) and \(j\). The uncertainty in \(l_{ij}\) is also incorporated, ensuring a consistent and biologically plausible pose.

\paragraph{Final 3D Pose Estimation (Node Level)}

Given the estimated 3D joint distributions
$P^{i} \sim \mathcal{N}(\mu_{P^i}, \Sigma_{P^i})$, the full human pose is refined by incorporating joint-coherence constraints and, when available, LiDAR-based 3D measurements. This refinement is formulated as a maximum likelihood estimation (MLE) problem. For numerical stability, the product of likelihood terms is expressed in the equivalent negative log-likelihood form:
\begin{equation}
\hat{\mathbf{P}}
=
\arg\min_{\{\hat{\mu}_{P^i}\}}
\lambda_c J_c
+
\lambda_b J_b
+
\lambda_l J_l ,
\label{eq:final_pose_estimation}
\end{equation}
where \(J_c\), \(J_b\), and \(J_l\) represent the camera-projection, joint-coherence, and LiDAR measurement terms, respectively.

The camera-projection term penalizes the deviation between the refined joint locations and the initial uncertainty-aware 3D estimates:
\begin{equation}
J_c =
\frac{1}{N_J}
\sum_{i=1}^{N_J}
(\hat{\mu}_{P^i}-\mu_{P^i})^{\top}
\Sigma_{P^i}^{-1}
(\hat{\mu}_{P^i}-\mu_{P^i}),
\end{equation}
where \(N_J\) is the number of valid joints. The joint-coherence term enforces anatomical consistency between adjacent joints:
\begin{equation}
J_b =
\frac{1}{|\mathcal{E}|}
\sum_{(i,j)\in\mathcal{E}}
\frac{
\left(
\|\hat{\mu}_{P^i}-\hat{\mu}_{P^j}\|-l_{ij}
\right)^2
}{
\sigma_{ij}^{2}
},
\end{equation}
where \(\mathcal{E}\) denotes the set of adjacent joint pairs, \(l_{ij}\) is the prior mean bone length between joints \(i\) and \(j\), and \(\sigma_{ij}\) characterizes the corresponding uncertainty.

When valid LiDAR associations are available, the LiDAR term provides additional 3D geometric constraints:
\begin{equation}
J_l =
\begin{cases}
\frac{1}{|\mathcal{L}|}
\sum_{k\in\mathcal{L}}
(\hat{\mu}_{P^k}-\mu_{L^k})^{\top}
\Sigma_{L^k}^{-1}
(\hat{\mu}_{P^k}-\mu_{L^k}),
& |\mathcal{L}|>0,\\
0,
& |\mathcal{L}|=0,
\end{cases}
\end{equation}
where \(\mathcal{L}\) is the subset of joints with valid LiDAR associations, \(\mu_{L^k}\) is the LiDAR-derived 3D measurement for joint \(k\), and \(\Sigma_{L^k}\) is its measurement covariance. The normalization by \(N_J\), \(|\mathcal{E}|\), and \(|\mathcal{L}|\) prevents any residual group from dominating solely due to the number of available terms.

The scalar weights \(\lambda_c\), \(\lambda_b\), and \(\lambda_l\) balance the relative contributions of the camera-projection, joint-coherence, and LiDAR terms. Since each residual is already weighted by its corresponding uncertainty, these scalar weights only regulate the relative importance of different information sources. The LiDAR term is activated only when valid LiDAR-to-joint associations are available. If LiDAR measurements are sparse or absent, \(\mathcal{L}\) becomes empty and \(J_l=0\), allowing the optimization to naturally fall back to the camera-only and joint-coherence formulation. In practice, LiDAR associations are accepted only after image-plane and world-frame consistency checks, which improves robustness against isolated or incorrectly associated LiDAR points.


\subsection{Delay-aware Global Perception}
In the ISN system, computational and communication delays can introduce inconsistencies in perception, severely degrading the performance of the subsequent motion-planning task.
To address this, our previous work proposed a delay-aware cooperative perception pipeline that effectively mitigates latency effects and provides consistent, real-time perception outputs \cite{ning2024enhancingindoormobilityconnected}. It contains three key components: global fusion across sensor nodes, multi-object tracking, and delay compensation.

Each ISN sends its local perception results to the CL, and the detections are matched based on spatial proximity and class labels. For objects detected by multiple ISNs, instead of simply averaging their positions like previous works \cite{ning2024enhancingindoormobilityconnected}, the CL refines the 3D human pose estimation via Eq. \ref{eq:final_pose_estimation} at the CL by fusing multi-view pose constraints from different ISNs to enhance robustness against occlusions and partial views.

Then, CL performs multi-object tracking using an extended Kalman filter (EKF) for each detected object to keep track of their states over time. Human trajectories, in particular, are modelled using a non-linear dynamic system capable of capturing changes in walking direction and speed, which enhances robustness in crowded and dynamic environments.

To further improve real-time responsiveness, the system incorporates timestamp-based delay compensation. Each ISN assigns a local timestamp to detected objects, allowing the CL module to compute the delay relative to its current time. Object positions are then extrapolated forward using motion prediction models, effectively aligning stale data with the current system state. This compensation mechanism ensures that the fused perception remains temporally consistent, even under variable network or processing latencies.

\section{Social-aware Motion Planning}
\begin{figure}[t]
    \centering
    \includegraphics[width=0.4\textwidth]{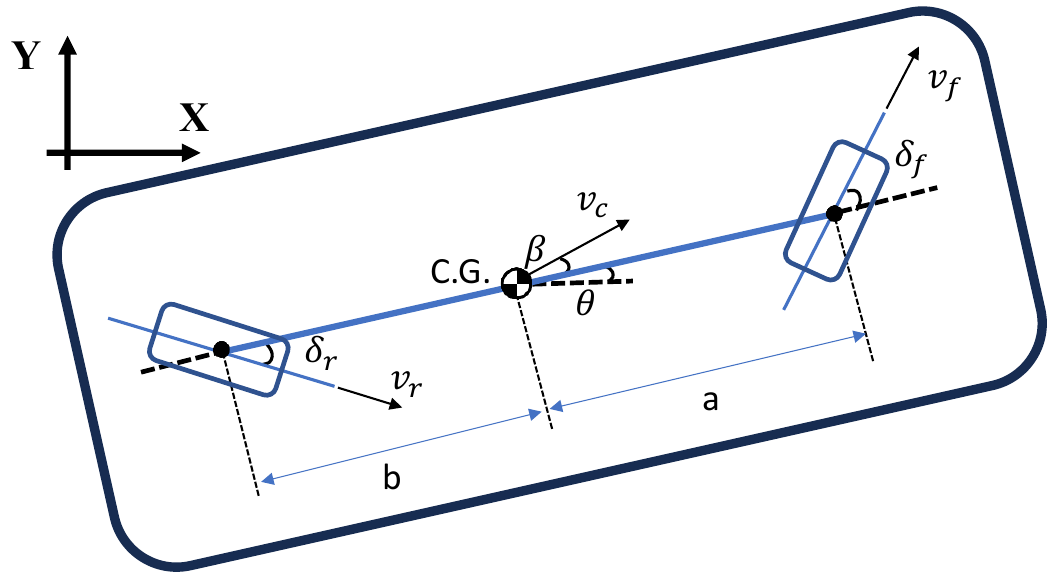}
    \vspace*{-0.1cm}
    \caption{Kinematic Motion Model of Robot}
    \label{fig:kin_model}
\end{figure}
\subsection{Kinematic-based Motion Model} 
A kinematic model is used to describe the robot's motion as it operates at speeds below 5 km/h. The model is based on the assumptions of tire non-slippage and rigid body dynamics.
Notably, the robot is capable of independently controlling the speed and steering angles of both the front and rear wheels, allowing it to have omnidirectional maneuverability. The kinematic model of the proposed robot is formulated as follows. 

\begin{gather}
    \beta =
    \tan^{-1}\left(
    \frac{
    l_r\tan(\delta_f)+l_f\tan(\delta_r)}
    {l_f+l_r}
    \right) \\
    v_c =
    \frac{
    v_f\cos(\delta_f)+v_r\cos(\delta_r)}
    {2\cos(\beta)} \\
    \dot{X}
    = v_c\cos(\psi+\beta) \quad
    \dot{Y}
    = v_c\sin(\psi+\beta) \\
    \dot{\psi}
    =
    \frac{
    v_f\sin(\delta_f)-v_r\sin(\delta_r)}
    {l_f+l_r}
\end{gather}

Variables $X$ and $Y$ represent the robot's position in the ground-fixed frame; $\psi$ is the robot's yaw angle; $\beta$ is the side-slip angle at the center of gravity (C.G.); $v_c$ is the C.G. velocity; $l_f$ and $l_r$ are the distances from the C.G. to the front and rear wheels; and $v_f$, $v_r$, $\delta_f$, and $\delta_r$ are the velocity and steering of the front and rear wheels, respectively. 
In this paper, the system states, $\zeta$, system outputs, $\eta$, and system control inputs, $u_c$ are defined as follows. 

\begin{equation}
    \zeta = \eta = \begin{bmatrix} X & Y & \psi \end{bmatrix}^\top
\end{equation}
\begin{equation}
    u_c = \begin{bmatrix} v_f & v_r & \delta_f & \delta_r \end{bmatrix}^\top
\end{equation}

\subsection{Personal Space Field}
\begin{figure} 
     \centering
     \begin{subfigure}[b]{0.25\textwidth}
         \centering
         \includegraphics[width=\textwidth]{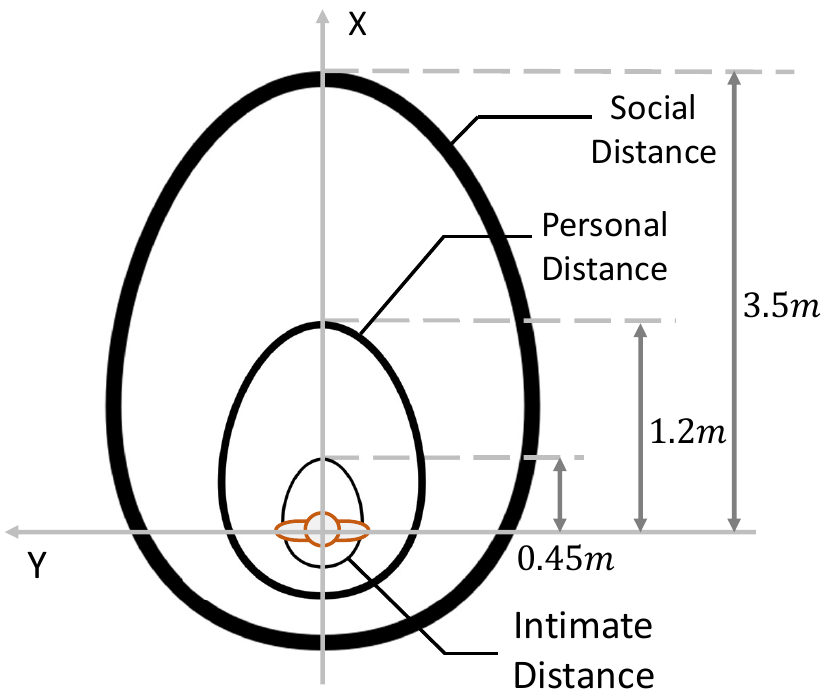}
         \caption{}
         \label{fig:y equals x}
     \end{subfigure}
     \hfill
     \begin{subfigure}[b]{0.21\textwidth}
         \centering
         \includegraphics[width=\textwidth]{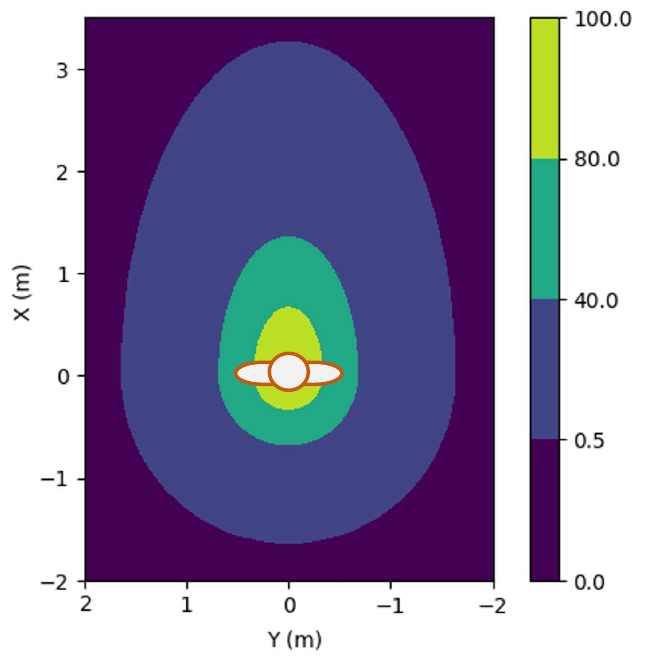}
         \caption{}
         \label{fig:three sin x}
     \end{subfigure}
        \caption{Graphical representation of the PS field: a) definition of Personal Field in psychology \cite{amaoka2009personal}. b) constructed a personal field using a mathematical expression. In both figures, $X$ and $Y$ represent the body-fixed coordinates of the human.}
        \label{fig:three graphs}
\end{figure}

\begin{table}[t]
\centering
\caption{Personal Space Distance Specifications \cite{amaoka2009personal}}
\label{tab:Personal Field Table}
  \begin{tabular}{lll} 
    \toprule
    \textbf{Types of Distance} & \textbf{Values} & \textbf{Relationship} \\
    \midrule
    Intimate distance      & 0 - 0.45 m    &Very intimate relationship\\
    Personal distance      & 0.45 - 1.2 m    &Friends\\
    Social distance      & 1.2 - 3.5 m    &Strangers\\
    Public distance      & $>$ 3.5 m      &Public speaking\\
    \bottomrule
  \end{tabular}
\end{table}
This paper adapts the concept of personal space (PS) \cite{amaoka2009personal} to model human behaviour in interaction with the robot. Compared with using the APF to represent the human, the PS model captures an individual's non-verbal and non-contact communication channels, influencing how they perceive proximity to others. Based on studies by Edward T. Hall \cite{hall1968proxemics} and R. Sommer \cite{sommer1969personal}, PS can be divided into four regions: public distance, social distance, personal distance, and intimate distance. These regions define the various zones of comfort and social interaction. The graphical representation and the relative distances can be found in Fig. \ref{fig:three graphs} a) and Table \ref{tab:Personal Field Table}. 

By assigning appropriate values to each region, as shown in Fig. \ref{fig:three graphs} b), the robot is able to maintain a psychologically safe distance from nearby humans. In this work, we adopt the asymmetric field proposed in \cite{amaoka2009personal}, where the field is intentionally shaped to be asymmetric along the human’s facing direction. This design reflects the natural psychological comfort zones observed in human interactions, where individuals are more sensitive to being disturbed from the front. As a result, the robot can anticipate and steer earlier to avoid potential discomfort.

In contrast, the conventional SFM and Artificial Potential Field (APF) are normally symmetrical in both $x$ and $y$ directions, which requires enlarging the overall field size to achieve a similar early-dodging behavior from the robot. However, because the conventional field covers an equal area in front of and behind the human, it can lead to inefficient navigation behavior, such as unnecessary deceleration or even being trapped when operating in densely populated environments. The proposed PS field, therefore, enables more natural and efficient avoidance while ensuring safe interpersonal distances.

    

For each detected human $q$, the robot position is first transformed into the local coordinate frame of the human. Let $\{X_{\mathrm{rob}},Y_{\mathrm{rob}}\}$ denote the global position of the robot, and let $\{X_q,Y_q,\theta_q\}$ denote the global position and heading direction of the $q$th human. The relative robot position in the human-centred coordinate frame is defined as
\begin{equation}
\mathbf{d}_q = \begin{bmatrix} 
X'_{{rob},q} \\ Y'_{{rob},q}
\end{bmatrix}= \begin{bmatrix}
\cos \theta_q & \sin \theta_q \\
-\sin \theta_q & \cos \theta_q
\end{bmatrix}
\begin{bmatrix}
X_{\mathrm{rob}} - X_q \\
Y_{\mathrm{rob}} - Y_q
\end{bmatrix}.
\label{equ:relative_position}
\end{equation}
where $X'_{{rob},q}$ represents the longitudinal relative position of the robot with respect to the heading direction of the (q)th human, while $Y'_{{rob},q}$ represents the lateral relative position.

To describe the directional asymmetry of human personal space, a binary function, $\gamma_q$, 
is introduced such that $\gamma_q=1$ when the robot is in front of the human ($X'_{\mathrm{rob},q}>0$) and $\gamma_q=0$ when it is behind the human ($X'_{\mathrm{rob},q}<0$). An estimation of such a function can be expressed as follows, where the variable $k_\gamma >0$ controls the steepness of the function. 

\begin{equation}
\gamma_q = 0.5(\tanh(k_\gamma\cdot X'_{{rob},q} / 2) + 1)
\label{equ:gamma}
\end{equation}

Furthermore, each front and rear personal space $\Omega_{q,i}, i\in\{f,r\}$ is defined as a 2D Gaussian function using the covariance matrix $\Sigma_i$ and relative distance $\mathbf{d}_q$. 
\begin{equation}
    \Omega_{q,i} = e^{-0.5\mathbf{d}_q^\top\Sigma_i^{-1}\mathbf{d}_q}, i\in\{f,r\}
\end{equation}

Here, $\Sigma_i$ defines the shape of the $\Omega_{q,i}$. Based on the studies of Shozo \cite{shozo1990comfortable}, it is suggested that the front PS, $\Omega_{q,f}$, of a human should be twice as large as the rear, left, and right sides. To implement this, the $\Sigma_f$ and $\Sigma_r$ are defined as follows, where $\sigma_{xx}=2\sigma_{yy}$. 

\begin{equation}
    \Sigma_f = 
    \begin{bmatrix} 
        \sigma_{xx} & 0 \\ 
        0 & \sigma_{yy} 
    \end{bmatrix}, \quad
    \Sigma_r = 
    \begin{bmatrix} 
        \sigma_{yy} & 0 \\ 
        0 & \sigma_{yy} 
    \end{bmatrix}
    \label{equ:personal_space_param}
\end{equation}

Amaoka et al. \cite{amaoka2009personal} stated that the values of $\sigma_{xx}$ should depend on several factors such as age, gender, etc.; however, since this paper is focused on the overall framework for an indoor mobility system, the values of $\sigma_{xx}$ are assumed to be fixed for all people when conducting experiments.

The final personal-space cost generated by the human $q$ is computed as follows, where $W_{\mathrm{ps}}$ is the weighting coefficient of the PS field.
\begin{equation}
\Omega_{q} = W_{\mathrm{ps}}(\gamma_q \Omega_{q,f}+(1-\gamma_q)\Omega_{q,r})
\end{equation}

\subsection{QP-Based MPC Approximation}

To achieve real-time computational efficiency, both the nonlinear kinematic model and the PS field are approximated before formulating the MPC problem. The nonlinear kinematic model is approximated using a first-order Taylor expansion around the current measured state and previous control input. Since the PS field depends on the relative position between the robot and the surrounding humans, the nominal trajectories of both the robot and humans need to be predicted over the prediction horizon. In this work, the nominal robot and human states are predicted using the Constant Velocity and Turning Rate (CVTR) model. The PS field is then approximated using a second-order Taylor expansion around these nominal predicted states. Specifically, for the robot position $z=[X,Y]^\top$, the nominal point $z_0$, and 
$\Delta z=z-z_0$, the PS field is approximated as
\begin{equation}
\begin{aligned}
\Omega(z) \approx\;& \Omega(z_0)
+ \nabla \Omega(z_0)^\top \Delta z \\
&+ \frac{1}{2}\Delta z^\top
\nabla^2 \Omega(z_0)\Delta z .
\end{aligned}
\label{eq:ps_second_order}
\end{equation}

Here, $z_0$ is updated at each prediction step according to the predicted robot state and the corresponding predicted human state. The constant term $\Omega(z_0)$ is omitted from the QP objective because it does not affect the optimizer. The gradient term contributes to the linear cost vector, while the Hessian term contributes to the quadratic cost matrix. To preserve the convexity of the QP, the PS-field Hessian is projected onto its positive semi-definite component before being included in the objective function. This quadratic approximation allows the PS-field cost to be incorporated into the QP formulation and efficiently solved using the OSQP solver \cite{osqp}. The analysis of the QP approximation of the PS field can be found in the supplementary material.

\subsection{Model Predictive Control-based Planner Design}
The proposed MPC consists of three objective terms and a set of constraints. Eq. \ref{equ:cost function} is the cost function of the MPC; Specifically, $J_1$ is the PS field, which penalizes the robot when it enters a human’s personal space, where, $N_{\mathrm{human}}$ denotes the number of humans in the environment and $\Omega_q$ represents the social cost associated with the $q$th human; $J_2$ penalizes the tracking error between the predicted output $\eta$ and $\eta_{\mathrm{ref}}$ throughout the prediction horizon, $N_{p}$, weighted by the positive definite matrix $Q$; $J_3$ regulates the front and rear wheels' steering action $\delta=\begin{bmatrix} \delta_f & \delta_r  \end{bmatrix}^\top$ and the deviation of the robot front and rear wheels' velocity $V = \begin{bmatrix} v_f & v_r  \end{bmatrix}^\top$ from its reference $V_{ref} = \begin{bmatrix} v_{f,ref} & v_{r,ref}  \end{bmatrix}^\top$ over the control horizon $N_{c}$, with weighting matrices $R$ and $G$, respectively.

Furthermore, Eq. \ref{equ:states} and \ref{equ:outputs} describe the motion model of the robot using the discretized linearized kinematic model, where $A_k$, $B_k$, and $C_k$ are the linearized system matrices at time step $k$, and $d_k$ represents model mismatch terms compared with the nonlinear model. Constraints are imposed on both the control increment and the absolute control input using  Eq. \ref{equ:input rate constraints} and  Eq. \ref{equ:input constraints}, where $\Delta u_{c\min}$ and $\Delta u_{c\max}$ denote the lower and upper bounds of control rate changes, and $u_{c\min}$ and $u_{c\max}$ define the control limits. In addition, the inequality Eq. \ref{equ:nonslip constraints} is a linearized wheel-speed constraint designed to prevent slippage due to speed differences between the front and rear wheels in the body-fixed frame, where $g(u_c) = |v_fcos(\delta_f)-v_rcos(\delta_r)|$ is the nonlinear wheel difference expression; $E(u_c)$ is the gradient of the function $g$ with respect to $u_c$. Without this constraint, the motion planner may generate control inputs that violate the non-slip assumption in the kinematic motion model, potentially leading to unreliable or unsafe robot behaviour. Additional details regarding this constraint can be found in \cite{yang2024intelligentmobilityintegratedmotion}. Finally, Eq. \ref{equ:initial cond.} specifies the initial condition of the optimization problem as the system states at the current time step.

\setlength{\abovedisplayskip}{1pt}
\setlength{\belowdisplayskip}{1pt}

\begin{equation}
u_c^{*}
=
\arg\min_{u_c}
J=J_1+J_2+J_3
\label{equ:cost function}
\end{equation}
\textit{where,}
\begin{gather}
J_1
=
\sum_{i=0}^{N_p-1}
\sum_{q=1}^{N_{\mathrm{human}}}
\Omega_q(z(k+i+1))
\\
J_2
=
\sum_{i=0}^{N_p-1}
\|\eta(k+i+1)-\eta_{\mathrm{ref}}\|_Q^2
\\
J_3
=
\sum_{j=0}^{N_c-1}
\left(
\|\delta(k+j)\|_R^2
+
\|V(k+j)-V_{\mathrm{ref}}\|_G^2
\right)
\end{gather}
\textit{s.t.}
\begin{gather}
\zeta(k+i+1)
=
A_k\zeta(k+i)
+
B_k u_c(k+j)
+
d_k(k+i)
\label{equ:states}\\
\eta(k+i+1)
=
C_k\zeta(k+i+1)
\label{equ:outputs}\\
\Delta u_{c,\min}
\leq
\Delta u_c(k+j)
\leq
\Delta u_{c,\max}
\label{equ:input rate constraints}\\
u_{c,\min}
\leq
u_c(k+j)
\leq
u_{c,\max}
\label{equ:input constraints}\\
\left|
E(u_c(0))u_c(k+j)
+
g(u_c(0))
\right|
\leq 0.1
\label{equ:nonslip constraints}\\
\zeta(k)=\zeta(0)
\label{equ:initial cond.}
\end{gather}

\section{Results}
This section presents the simulation and experimental results used to evaluate the effectiveness and practicality of the proposed SAP-CoPE framework. First, the perception system is validated by comparing its detection accuracy with baseline methods under computation and communication delays. Next, the proposed social-aware motion planning algorithm is evaluated in a crowded open-space simulation. Finally, the overall system is assessed in scenarios with varying numbers of humans, focusing on motion planning performance and comparisons with other planning algorithms. 
\subsection{Perception Experiments}
\subsubsection{Evaluation Metrics}
To rigorously evaluate the performance of the proposed 3D human pose estimation method, we use two principal metrics, compared against annotated ground truth data: Mean Absolute Error (MAE) and Root Mean Square Error (RMSE). The MAE reflects the average magnitude of estimation errors, while RMSE gives greater weight to large errors. The evaluation metrics specifically include  \textbf{position error} (measured in meters) and  \textbf{yaw error} (measured in degrees for orientation).

\begin{figure*}[!t]
    \centering
    \includegraphics[width=\textwidth]{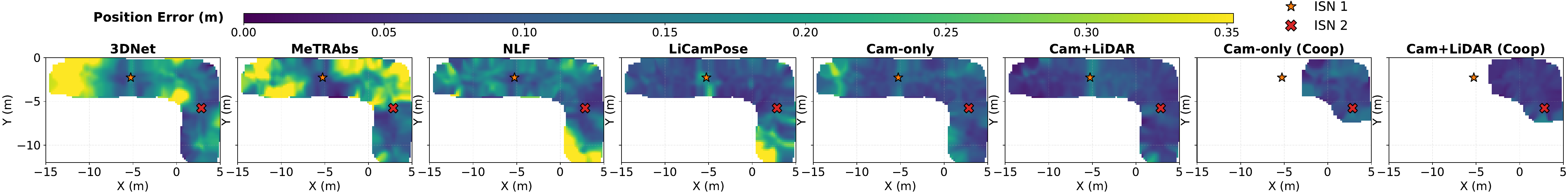}
    \caption{Spatial distribution of position errors.}
    \label{fig:position_error_dist}
\end{figure*}
\begin{figure*}[!t]
    \centering
    \includegraphics[width=\textwidth]{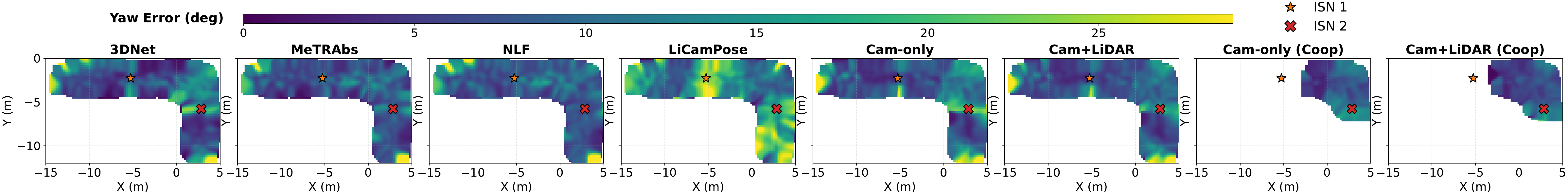}
    \caption{Spatial distribution of yaw errors.}
    \label{fig:yaw_error_dist}
\end{figure*}

\begin{table}[t]
    \centering
    \caption{Pose Estimation Performance Evaluation}
    \label{tab:pose_evaluation}
    \begin{tabular}{l|c|c}
        \hline
        \textbf{Method} & \textbf{Position Error (m)} & \textbf{Yaw Error (deg)} \\
                        & MAE / RMSE                  & MAE / RMSE               \\
        \hline
        3DNet~\cite{martinez2017simple} (2017)       & 0.1795 / 0.2163 & 8.5629 / 12.1001 \\
        Metrabs~\cite{sarandi2023learning} (2023)    & 0.2294 / 0.2926 & 8.1819 / 11.2037 \\
        NLF~\cite{sarandi2024nlf} (2024)              & 0.1659 / 0.2103 & 8.6010 / 11.3837 \\
        LiCamPose~\cite{pan2025licampose} (2025)      & 0.1188 / 0.1653 & 15.0739 / 19.3666 \\
        Cam-only                               & 0.1098 / 0.1308 & 8.5199 / 11.7416 \\
        Cam+LiDAR                              & 0.0892 / 0.1090 & 7.9168 / 11.0136 \\
        Cam-only (coop)                        & 0.1087 / 0.1237 & 8.1727 / 10.2264 \\
        Cam+LiDAR (coop)                       & 0.0771 / 0.0862 & 6.8607 / 8.9542  \\
        \hline
    \end{tabular}
    \vspace{1mm}
    
    \begin{minipage}{0.48\textwidth}
    \footnotesize
    \textit{Note:} 3DNet, Metrabs, and NLF are evaluated using their officially released pre-trained models. Since LiCamPose does not provide a directly usable pre-trained model, it is trained following the official repository instructions. For the proposed methods, no manual human-pose annotations are needed on ICID; human 2D pose labels are generated by YOLOv11-x, while only the autonomous bed is manually annotated with 2D bounding boxes. The resulting ICID annotations are co-trained with COCO to train the YOLOv11-s model used in our perception pipeline.
    \end{minipage}
\end{table}



\begin{table}[t]
\centering
\scriptsize
\caption{Delay mitigation evaluation results}
\begin{tabular}{@{}cccccc@{}}
\toprule
\textbf{Delay} & \textbf{Mode} & \textbf{F1} & \textbf{Recall} & \textbf{Avg. DE} & \textbf{Avg. AE} \\
\midrule
\multirow{2}{*}{0ms} & Baseline & 0.9456 & 0.9512 & 0.0829 & 7.2475 \\
 & Delay-Aware & \textbf{0.9485} & \textbf{0.9739} & \textbf{0.0818} & \textbf{5.9557} \\
\midrule
\multirow{2}{*}{50ms} & Baseline & 0.9456 & 0.9510 & 0.0864 & 7.2314 \\
 & Delay-Aware & \textbf{0.9480} & \textbf{0.9739} & \textbf{0.0840} & \textbf{6.1045} \\
\midrule
\multirow{2}{*}{100ms} & Baseline & 0.9455 & 0.9510 & 0.0975 & 7.2110 \\
 & Delay-Aware & \textbf{0.9480} & \textbf{0.9738} & \textbf{0.0876} & \textbf{6.2546} \\
\midrule
\multirow{2}{*}{200ms} & Baseline & 0.9449 & 0.9513 & 0.1439 & 7.2365 \\
 & Delay-Aware & \textbf{0.9479} & \textbf{0.9740} & \textbf{0.0993} & \textbf{6.5910} \\
\bottomrule
\end{tabular}
\label{tab:eva_result_delay}
\end{table}

\subsubsection{Results}

We evaluate both pose estimation accuracy and system robustness under computation and communication delay. 
For pose estimation, we compare eight configurations: learning-based baselines 
(3DNet\cite{martinez2017simple}, Metrabs\cite{sarandi2023learning}, NLF\cite{sarandi2024nlf}, LiCamPose\cite{pan2025licampose}), single-node camera-only and camera--LiDAR fusion, and their 
cooperative multi-node counterparts. Quantitative metrics are shown in 
Table~\ref{tab:pose_evaluation}, and spatial error distributions are visualized in 
Figs.~\ref{fig:position_error_dist} and \ref{fig:yaw_error_dist}.

\textbf{Position error:}
Among the image-based learning baselines, NLF achieves the lowest position error, with 0.1659\,m MAE and 0.2103\,m RMSE, outperforming both 3DNet and Metrabs. LiCamPose further reduces the position error to 0.1188\,m MAE by incorporating camera and LiDAR information. Nevertheless, the proposed geometry-based methods still achieve lower localization errors. Even the single-node Cam-only variant obtains 0.1098\,m MAE, while adding sparse LiDAR measurements reduces the MAE from 0.1098\,m to 0.0892\,m. When cooperative observations are available in the overlapping sensing region, performance further improves: Cam+LiDAR (coop) achieves the lowest error of 0.0771\,m MAE and 0.0862\,m RMSE, outperforming all evaluated baselines. The advantages of the proposed method are also reflected in the spatial heatmaps in Fig.~\ref{fig:position_error_dist}. The learning-based methods show larger errors in distant or low-observability regions, while the proposed methods maintain more consistent accuracy. This improvement comes from the uncertainty-aware pose estimation formulation, which explicitly models the camera projection process and propagates measurement uncertainty through the geometry.

\textbf{Yaw error:}
Orientation estimation shows a similar trend, although the improvements are more moderate. Among the learning-based baselines, Metrabs obtains the lowest yaw MAE, while NLF achieves comparable performance. LiCamPose improves position estimation compared with image-only learning baselines, but its yaw error is higher in our top-down infrastructure setting, indicating that orientation estimation remains challenging when adapting multi-modal learning-based methods to this deployment. In contrast, the proposed methods provide more stable heading estimation. The Cam+LiDAR variant reduces yaw MAE to 7.92$^{\circ}$, and the cooperative Cam+LiDAR variant further reduces it to 6.86$^{\circ}$, achieving the best overall orientation accuracy. This demonstrates that the proposed uncertainty-aware geometric formulation and cooperative observations help stabilize heading estimation under occlusions and ambiguous top-down views.

\subsubsection{Delay Robustness Evaluation}
In our system, local perception requires an average of 42\,ms, while communication latency is approximately 3\,ms over local Wi-Fi and 53\,ms over 5G; global fusion and tracking add a further 2\,ms~\cite{ning2024enhancingindoormobilityconnected}.
Consequently, practical deployments may experience total latency ranging from tens to hundreds of milliseconds. To quantify robustness to latency, we evaluate performance under simulated communication delays from 0 to 200\,ms, with results summarized in Table~\ref{tab:eva_result_delay}.

The baseline method directly fuses all received observations without multi-object tracking and delay compensation, while the proposed delay-aware pipeline incorporates timestamp-based delay correction and motion prediction.
Across all delay settings, the delay-aware pipeline preserves stable perception quality and consistently outperforms the baseline, yielding higher recall as well as lower displacement error (Avg.\ DE) and angular error (Avg.\ AE). The advantage becomes more obvious as delays increase: at 200\,ms, the baseline position error rises to 0.1439\,m, whereas the delay-aware approach limits it to 0.0993\,m. Similarly, the yaw error decreases from 7.24$^{\circ}$ to 6.59$^{\circ}$. These results indicate that timestamp-based compensation coupled with motion prediction effectively mitigates stale observations and improves temporal consistency under significant communication latency.





Overall, the perception results show that sparse LiDAR fusion, cooperative multi-node observations, and delay-aware compensation jointly improve the accuracy and temporal consistency of human pose estimation, providing reliable perception inputs for the subsequent social-aware motion-planning evaluation.

\subsection{Motion Planning Simulation}
\begin{table*}[!t]
    \centering
    \caption{Simulation Statistics Results Comparison in the Dense Crowd Scenario}
    \label{tab:performance_metric_dense}
    \begin{tabular}{|c|c|c|c|c|c|c|}
    \hline
    \makecell{Number of\\Human} 
    & Method 
    & \makecell{Collision / Stuck Rate $\downarrow$\\{[\%]}} 
    & \makecell{Clearance $\uparrow$\\(Avg. / Std.) [m]} 
    & \makecell{Total Traveled Dist. $\downarrow$\\(Avg. / Std.) [m]} 
    & \makecell{Total Traveled Time $\downarrow$\\(Avg. / Std.) [sec]} 
    & \makecell{Computational Time $\downarrow$\\(Avg. / Std.) [ms]} \\
    \hline
    \multirow{4}{*}{15}
    & SFM & 50 / 4 & 0.10 / 0.025\textsuperscript{a} & 18.02 / 0.76\textsuperscript{a} & 25.05 / 7.19\textsuperscript{a} & \textbf{8.56 / 1.55} \\
    & SARL & 50 / 0 & 0.12 / 0.011\textsuperscript{a} & 18.79 / 2.47\textsuperscript{a} & 22.91 / 3.62\textsuperscript{b} & 29.66 / 2.38 \\
    & MPC + PF & 10 / 3 & \textbf{0.18 / 0.022\textsuperscript{b}} & 16.61 / 0.30\textsuperscript{b} & 23.31 / 4.82\textsuperscript{ab} & 37.69 / 3.31 \\
    & Proposed & \textbf{7 / 0} & 0.17 / 0.090\textsuperscript{b} & \textbf{16.50 / 0.24\textsuperscript{c}} & \textbf{20.36 / 3.67\textsuperscript{c}} & 41.69 / 4.72 \\
    \hline
    \end{tabular}
\end{table*}

To evaluate the performance of the proposed motion planner in a larger and crowded scenario, simulations were conducted and compared with three baseline methods. The first baseline is the conventional Social Force Model (SFM) \cite{helbing1995social}, which was modified to account for the rectangular shape of the robot. The second baseline is MPC with a symmetric potential field (MPC+PF)~\cite{rasekhipour2016potential}, which is used to evaluate the benefit of replacing a conventional symmetric field representation with the proposed asymmetric PS field. The third baseline is SARL~\cite{chen2019crowd}, a deep reinforcement learning-based social-aware navigation method.

The simulation environment was designed as a $12\mathrm{m} \times 12\mathrm{m}$ open square space to represent a hospital lobby-like scenario. A total of 15 pedestrians were included, and each pedestrian was modelled using SFM to simulate actual human motion. Each method was evaluated over 100 trials, with randomized initial and goal positions for the pedestrians. The same random seeds were used across all methods for each trial to ensure a fair comparison. 

Four evaluation metrics were used to assess the performance of the proposed motion planning algorithm against other baselines. 
a) \textbf{\textit{Collision and Stuck Rates}}: this metric evaluates the task-completion performance of each method. A successful trial is defined as one in which the robot reaches the goal without collision or becoming stuck. A collision case refers to physical contact between the robot and a human (boundary-to-
boundary distance), while a stuck case refers to the robot failing to reach the goal within the allowed time due to local minima, excessive stopping, or infeasible motion. 
b) \textbf{\textit{Clearance}}: this metric measures the minimum boundary-to-boundary distance between the robot and the human. Lower values indicate that the robot moves closer to the human, which may increase psychological discomfort and collision risk. 
c) \textbf{\textit{Travel Distance}}: this metric measures the total traveled distance of the robot from the starting position to the goal, reflecting travel efficiency. 
d) \textbf{\textit{Travel Time}}: this metric measures the total traveling time from the starting position to the goal, reflecting motion efficiency and smoothness. A shorter travel time indicates that the robot can maintain the desired speed with fewer unnecessary decelerations or stops. 

Table \ref{tab:performance_metric_dense} shows the simulation results. Statistical significance among the methods was evaluated using the Wilcoxon signed-rank test (a paired, non-parametric test) with Holm–Bonferroni correction for multiple comparisons. The resulting groupings are indicated by superscript letters in Table \ref{tab:performance_metric_dense}. For each performance metric, methods that share at least one superscript letter are not significantly different, whereas methods with no letter in common are significantly different. Since failed trials may yield incomplete or non-comparable path and time measurements, only successful trials were used in the statistical comparison of total traveled distance and total traveled time.

Compared with the baseline methods, the proposed method achieved the best overall success performance, reducing the collision rate to 7\% and eliminating stuck cases. In contrast, both SFM and SARL resulted in a 50\% collision rate, while MPC+PF produced a 10\% collision rate and a 3\% stuck rate. These results indicate that the proposed method provides improved robustness in dense human-populated environments, including when compared with the evaluated learning-based SARL baseline.

In terms of safety-related performance, the proposed method maintained a clearance distance statistically comparable to MPC+PF, while both methods significantly outperformed SFM and SARL. Although MPC+PF achieved a slightly higher average clearance, its higher collision/stuck rate and longer travel time indicate that increased clearance alone does not necessarily lead to better overall navigation performance. For navigation efficiency, the proposed method achieved the shortest average traveled distance and traversal time among all evaluated methods. The statistical results show that the proposed method significantly reduced the traveled distance compared with all baselines. In addition, based on the post-hoc statistical grouping, the proposed method achieved significantly shorter travel time than the other evaluated methods. 

To improve computational efficiency in dense environments, only pedestrians within a 5 m radius of the robot are considered when constructing the MPC optimization problem. This local-neighbour selection strategy limits the number of active pedestrian-related terms while retaining the most relevant agents for collision avoidance and social interaction. In the 15-pedestrian dense-crowd simulation, the proposed method achieved an average computation time of 41.69 ms, which remains below the MPC sampling period of 100 ms. Although higher than the baseline methods, this result demonstrates that the proposed formulation remains computationally feasible under the tested dense-crowd condition.

Overall, the dense-crowd simulation results demonstrate that the proposed method provides a favourable trade-off among safety, robustness, navigation efficiency, and computational feasibility. Some simulation quantitative results can be found in the repository. 

\subsection{Combined Experiments}
Since the perception system has been validated in the previous section, this section focuses on motion planning using the proposed perception outputs. The perception results with computation and communication delay mitigation are sent to all motion planners, including the baseline planners. The experiments were conducted in a hallway-like environment with a straight and a 90-degree turn setup. During the experiments, the robot was required to follow the predefined reference trajectory as closely as possible while proactively avoiding collisions. 

\subsubsection{Mobile Robot}
The robot is retrofitted based on an existing medical bed. A picture of the robot is shown in Fig. \ref{fig:framework}, in the "Application Layer" block. To ensure optimal traction and handling, the robot incorporates a two-wheel-drive system (one wheel in the front and one in the back) and an independent steering system, allowing each wheel to turn independently for omnidirectional mobility. 

\subsubsection{Experimental Setup}
\begin{figure}[t]
    \centering
    \includegraphics[width=0.5\textwidth]{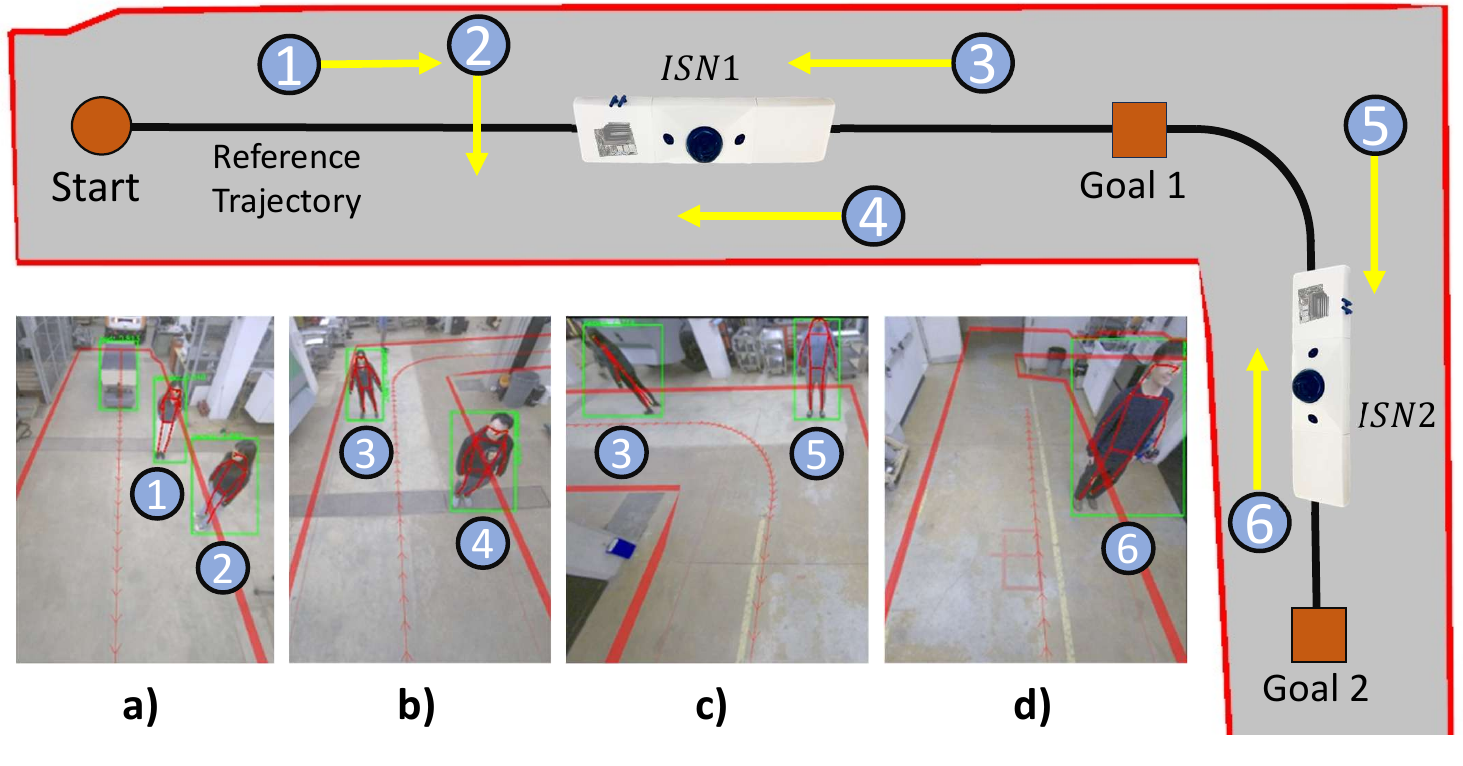}
    \vspace*{-0.8cm}
    \caption{Experiment setup with all participants. Images captured from a) ISN1's left camera. b) ISN1's right camera. c) ISN2's left camera. d) ISN2's right camera.}
    \label{fig: Experiment Setup}
\end{figure}

\begin{table}[t]
\centering
\caption{Robot and MPC Parameters. Units for velocity (or acceleration) are in [m/s] (or [m/s$^2$]), and steering angles (or steering rate) are in [rad] (or [rad/s]).}
\label{tab:Robot_and_MPC_Param}
  \begin{tabular}{cccc} 
    \toprule
    Symbol & Value & Symbol & Value \\
    \midrule
    $l_f, l_r$     & 1.2 m    & $u_{c\max}$     & $\left[0.7, 0.7, \frac{\pi}{2}, \frac{\pi}{2}\right]$\\
     $T_s$    & 0.1 sec    &$u_{c\min}$    & -$\left[0.7, 0.7, \frac{\pi}{2}, \frac{\pi}{2}\right]$\\
    $N_p$        & $20$    &$\Delta u_{c\max}$  & $\left[1, 1, \frac{\pi}{24}, \frac{\pi}{24}\right]$\\
    $N_c$      & 10   &$\Delta u_{c\min}$  & -$\left[1, 1, \frac{\pi}{24}, \frac{\pi}{24}\right]$ \\

    \bottomrule
  \end{tabular}
\end{table}

\begin{table*}[!t]
    \centering
    \caption{Experimental Statistics Results Comparison between the Proposed Method and Other Methods}
    \label{tab:performance_metric_planning}
    \footnotesize
    \setlength{\tabcolsep}{4pt}
    \renewcommand{\arraystretch}{1.15}
    \begin{tabular}{|c|ccc|ccc|ccc|}
    \hline
     & \multicolumn{3}{c|}{Clearance (Avg. / Std.) [m] $\uparrow$} & \multicolumn{3}{c|}{Traveled Distance (Avg. / Std.) [m] $\downarrow$} & \multicolumn{3}{c|}{Traveled Time (Avg. / Std.) [sec.] $\downarrow$} \\
    \hline
    \# Human & SFM & MPC + PF & Proposed & SFM & MPC + PF & Proposed & SFM & MPC + PF & Proposed \\
    \hline
    2& 0.22 / 0.056\textsuperscript{a}& \textbf{0.39 / 0.052\textsuperscript{b}}& 0.39 / 0.068\textsuperscript{b}& \textbf{12.05 / 0.15\textsuperscript{a}}& 12.23 / 0.078\textsuperscript{b}& 12.14 / 0.22\textsuperscript{ab}& 18.11 / 0.18\textsuperscript{a}& 17.78 / 0.15\textsuperscript{b}& \textbf{17.77 / 0.16\textsuperscript{b}}\\
    4& 0.071 / 0.048\textsuperscript{a}& 0.17 / 0.11\textsuperscript{b}& \textbf{0.29 / 0.048\textsuperscript{c}}& \textbf{12.06 / 0.11\textsuperscript{a}}& 12.21 / 0.23\textsuperscript{ab}& 12.33 / 0.19\textsuperscript{b}& 18.58 / 0.43\textsuperscript{a}& 19.74 / 2.31\textsuperscript{a}& \textbf{18.16 / 0.14\textsuperscript{b}}\\
    6& 0.022 / 0.031\textsuperscript{a}& 0.11 / 0.087\textsuperscript{b}& \textbf{0.26 / 0.042\textsuperscript{c}}& 21.69 / 1.06\textsuperscript{ab}& 21.86 / 0.59\textsuperscript{a}& \textbf{20.76 / 0.18\textsuperscript{b}}& 36.12 / 3.96\textsuperscript{a}& 33.43 / 1.58\textsuperscript{a}& \textbf{30.73 / 0.39\textsuperscript{b}}\\
    \hline
    \end{tabular}
\end{table*}

The proposed perception system and planner were implemented and tested in experiments using the C++ and Python platforms. To verify our proposed planner, the generated control inputs were directly sent to the robot for simplicity. The parameters for the robot and the MPC motion planner are summarized in Table \ref{tab:Robot_and_MPC_Param}. All computations in the CL, including global perception fusion and motion planning algorithms, are carried out on a Linux-based desktop equipped with an Intel i7-14700F CPU and 64GB of RAM.

The participant's initial position (numbered blue circles), movement direction (yellow arrow), and ISNs' location can be found in Fig. \ref{fig: Experiment Setup}. Three experimental scenarios were set up to validate the proposed system. In scenario (a), which involved 2 participants, the robot was asked to follow the reference path and stop at "Goal 1", while participants 3 and 4 were involved. In scenario (b), which involved 4 participants, the goal of the robot and the movement of participants 3 and 4 remained the same, but with the addition of participants 1 and 2. In scenario (c), which involved all participants, the goal of the robot was set to "goal 2", and participants 1 to 4 remained the same, but with the addition of participants 5 and 6.


\subsubsection{Evaluation Metrics and Baselines}
The experimental evaluation follows the same metric definitions used in the simulation study, but only clearance, traveled distance, and traveled time are reported. The collision and stuck rates are not included because the experiments were conducted with a limited number of controlled trials and under human-supervised safety conditions. Similarly, the experimental baselines are limited to SFM and MPC+PF, which are the classical model-based and force-field-based methods already evaluated in simulation. The SARL-based method was not included in the real-world experiments because its deployment would require additional integration, tuning, and safety validation on the physical robot platform, which is beyond the scope of this work. 


\subsubsection{Results}

\begin{figure}
    \centering
    \begin{subfigure}[b]{0.5\textwidth}
        \centering
        \includegraphics[width=\textwidth]{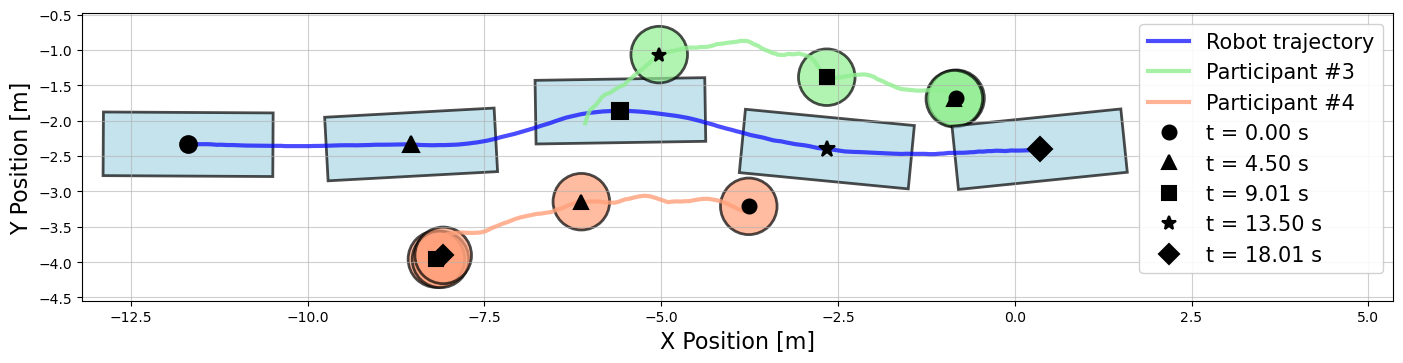}
        \caption{SFM}
        \label{fig:2Human_baseline_SFM}
    \end{subfigure}
    
    \begin{subfigure}[b]{0.5\textwidth}
        \centering
        \includegraphics[width=\textwidth]{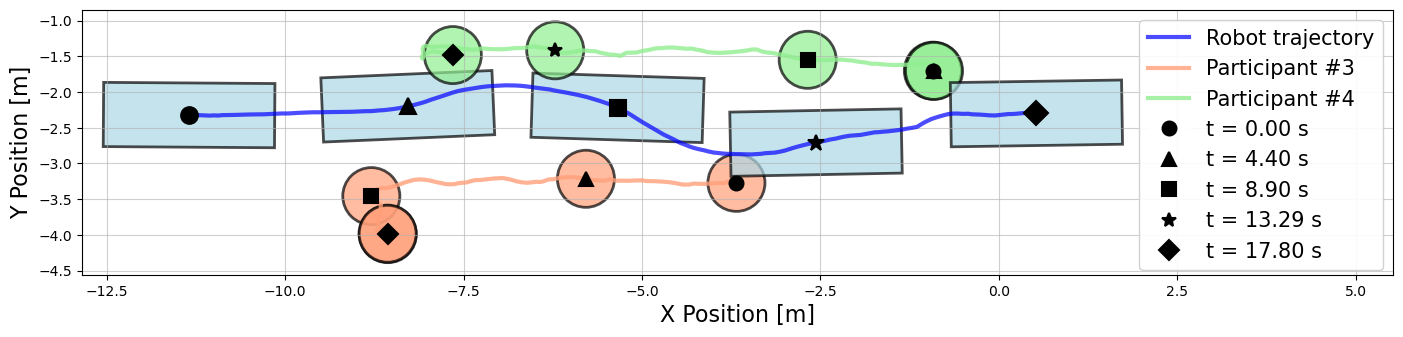}
        \caption{Proposed Method}
        \label{fig:2Human_proposed}
    \end{subfigure}
    \caption{Representative trajectories in the 2-participant experiment: (a) SFM; (b) proposed method combining MPC and the PS model.}
    \label{fig:2Human_experiment}
\end{figure}

\begin{figure}[t]
    \centering
    \includegraphics[width=0.5\textwidth]{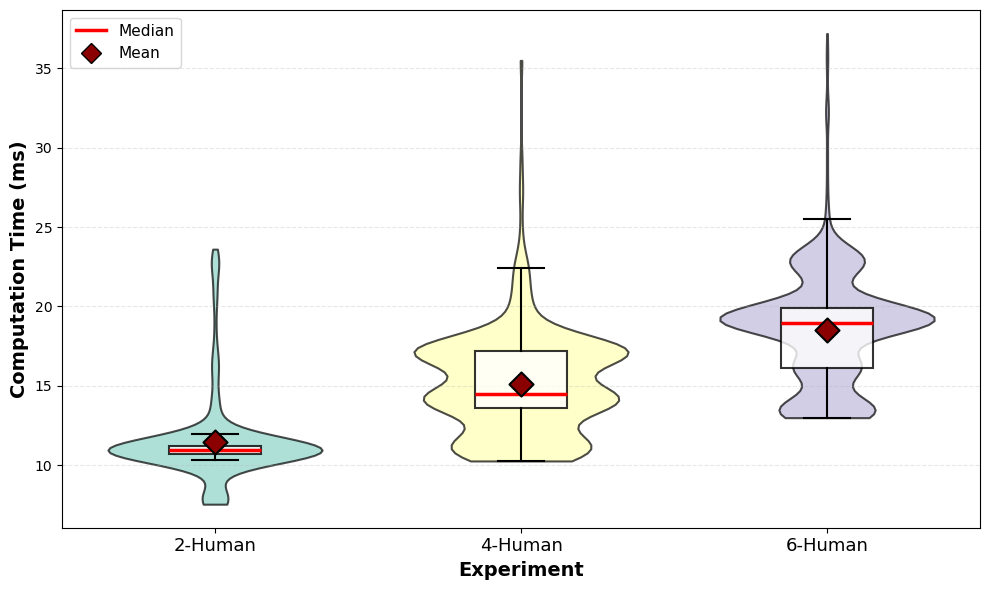}
    \vspace*{-0.5cm}
    \caption{Proposed Motion Planning Method Computation Time Distribution}
    \label{fig: computation_plot}
\end{figure}

Each method in each scenario was repeated 10 times to reduce the effect of randomness. The experimental results in Table \ref{tab:performance_metric_planning} show that the proposed method achieved the best overall performance across the tested scenarios. Statistical significance among the methods was evaluated similarly to the simulation section using the Wilcoxon signed-rank test, and the resulting groupings are indicated by superscript letters. Methods that share at least one superscript letter are not significantly different.

For clearance, the proposed method maintained a larger distance from participants than SFM in all scenarios. In the 2-participant scenario, the proposed method achieved the same average clearance as MPC+PF, and both methods were statistically comparable. Fig. \ref{fig:2Human_experiment} demonstrates the trajectory of the robot and human for SFM and the proposed method. The robot with SFM does not leave enough space when encountering participants, and they need to dodge the robot instead. In the 4- and 6-participant scenarios, the proposed method achieved the highest average clearance, with statistically significant improvements over both SFM and MPC+PF. As the number of participants increased, the clearance of the baseline methods decreased noticeably, particularly for SFM, whereas the proposed method maintained a larger safety margin. This result indicates that explicitly incorporating the PS field into the MPC formulation improves the planner's ability to preserve safe interaction distances in more crowded settings.

For navigation efficiency, the proposed method achieved the lowest average travel time in all three experimental scenarios. In the 2-participant scenario, the proposed method and MPC+PF showed statistically comparable travel times, while both were faster than SFM. In the 4- and 6-participant scenarios, the proposed method achieved significantly shorter travel times than the baseline methods. The advantage became more evident in the 6-participant case, where the proposed method reduced the travel time to 30.73 s, compared with 36.12 s for SFM and 33.43 s for MPC+PF. The traveled-distance results are more scenario-dependent. In the 2- and 4-participant cases, SFM produced the shortest average traveled distance, although the differences were not always statistically significant. In the 6-participant case, the proposed method achieved the shortest average traveled distance, reducing the distance to 20.76 m compared with 21.69 m for SFM and 21.86 m for MPC+PF. This suggests that the proposed method may select slightly longer paths in sparse settings to maintain comfortable clearance, while in denser settings, its social-aware planning strategy can improve both safety and navigation efficiency. Some experiment videos can be found in the repository. 

Fig. \ref{fig: computation_plot} further evaluates the computational performance of the proposed planner in the physical experiments using violin plots, combined with box plot elements. The computation time increased with the number of dynamic obstacles, from a mean of 11.5 ms in the 2-participant scenario to 15.0 ms in the 4-participant scenario and 18.7 ms in the 6-participant scenario. Occasional outliers were observed in the 4- and 6-participant cases, with the maximum computation time approaching approximately 37 ms, reflecting the increased optimization complexity caused by dense multi-obstacle interactions. Nevertheless, all computation times remained well below the MPC sampling period of 100 ms, confirming the real-time feasibility of the proposed planner in the tested experimental scenarios.

\subsection{Discussion}
\subsubsection{Motion Planning Simulation vs Experiment}
The experimental observations are generally consistent with the simulation results, particularly in clearance and travel time. In both cases, the proposed method maintained strong clearance performance while achieving the lowest average travel time in most scenarios. The traveled-distance results are more scenario-dependent: although the simulation showed a clearer advantage for the proposed method, the experiments showed statistically similar or mixed trends in the 2- and 4-participant cases, with the shortest distance achieved by the proposed method only in the 6-participant case. This difference is reasonable due to the limited experimental trials and sparse human density.

In summary, SFM showed the weakest safety performance because it does not explicitly consider the robot’s omnidirectional maneuverability when generating avoidance behaviour. MPC+PF performed competitively in sparse scenarios; however, its symmetric field representation can lead to conservative deceleration and reduced efficiency in denser settings. By contrast, the proposed method maintained a better balance between interaction-aware clearance, travel efficiency, and real-time computational feasibility, demonstrating its suitability for social-aware navigation in human-populated indoor environments.
\subsubsection{Component-wise Ablation Discussion}
The conducted evaluations provide component-wise evidence for the contribution of the main modules in SAP-CoPE. First, the contribution of LiDAR fusion is isolated in Table~\ref{tab:pose_evaluation} by comparing the Cam-only and Cam+LiDAR variants. Adding sparse LiDAR measurements reduces the position MAE from 0.1098\,m to 0.0892\,m in the single-node case, while the cooperative Cam+LiDAR configuration further reduces the error to 0.0771\,m. This confirms that LiDAR fusion improves spatial localization accuracy by providing additional depth constraints. 

Second, the effect of delay compensation is evaluated in Table~\ref{tab:eva_result_delay}, where the delay-aware pipeline is compared with the baseline fusion method under different artificial delays. As the delay increases to 200\,ms, the baseline displacement error increases to 0.1439\,m, whereas the delay-aware method limits the error to 0.0993\,m, demonstrating that timestamp-based correction and motion prediction improve temporal consistency under communication latency. 

Third, the contribution of the asymmetric PS field is isolated by comparing the proposed method with MPC+PF, which uses a symmetric potential field formulation. In both the dense-crowd simulation and the physical experiments, the proposed method achieves lower failure rates and improved navigation efficiency compared with MPC+PF. Specifically, the proposed method achieves the lowest collision/stuck rate, and the shortest average traveled distance and time in the dense-crowd simulation, while maintaining comparable minimum clearance. In the physical experiments, it also achieves the lowest average travel time across all tested participant numbers while maintaining safe clearance. These results indicate that the asymmetric PS field provides a more direction-aware representation of human personal space than the symmetric potential field, improving socially compliant avoidance behaviour without sacrificing navigation efficiency or real-time feasibility.


\section{Conclusions and Future Work}

In this paper, we have proposed SAP-CoPE, a social-aware navigation framework for autonomous mobile robots operating in human-populated indoor environments. The framework has integrated cooperative infrastructure-based perception with MPC-based social-aware motion planning. In the perception module, an uncertainty-aware 3D human pose estimation method has been developed to estimate human pose and heading from camera observations, with optional sparse LiDAR-based geometric constraints. In the planning module, the estimated human pose and orientation have been used to construct an asymmetric PS field, enabling the robot to generate trajectories that balance physical safety, human comfort, travel efficiency, and kinematic feasibility.

The proposed framework has been evaluated through perception experiments, dense-crowd simulation, and real-world robot experiments with human participants. The results have demonstrated improved pose estimation accuracy, robust delay-aware perception, and social-aware navigation performance compared with baseline methods. The computation-time results have also shown that the proposed planner has satisfied the real-time requirement under the tested indoor scenarios. These results have indicated that SAP-CoPE is suitable for infrastructure-supported indoor public environments where sensor nodes can provide cooperative perception.

Several limitations have been identified in this work. First, the physical experiments have been conducted in controlled hallway-like scenarios; therefore, larger-scale validation in more complex and dynamic environments, such as hospital wards or open lobbies, is still needed. Second, the current prior-height-based pose initialization has assumed approximately upright pedestrians, which is reasonable for the walking scenarios considered in this study but may be less accurate for sitting, bending, or other unusual postures. Finally, the CVTR method has been used to predict the robot and human poses as nominal points for the quadratic approximation of the PS field. However, sudden changes in pedestrian motion may introduce a mismatch between the predicted and actual PS fields, causing a potential collision. Future work will address these limitations by extending validation to more complex real-world environments and incorporating learning-based human intention and motion prediction to improve robustness under dynamic human motion.

\section*{Reproducibility Statement}
To support reproducibility for this study, we provide a repository at: \url{https://github.com/FrankYang248/SAP-CoPE-Project}. The repository includes selected implementation components, supplementary materials, and representative simulation and experiment results related to the proposed SAP-CoPE framework.

\bibliographystyle{IEEEtran}
\bibliography{ref}



\ifCLASSOPTIONcaptionsoff
  \newpage
\fi



\begin{IEEEbiography}[{\includegraphics[width=1in,height=1.25in,clip,keepaspectratio]{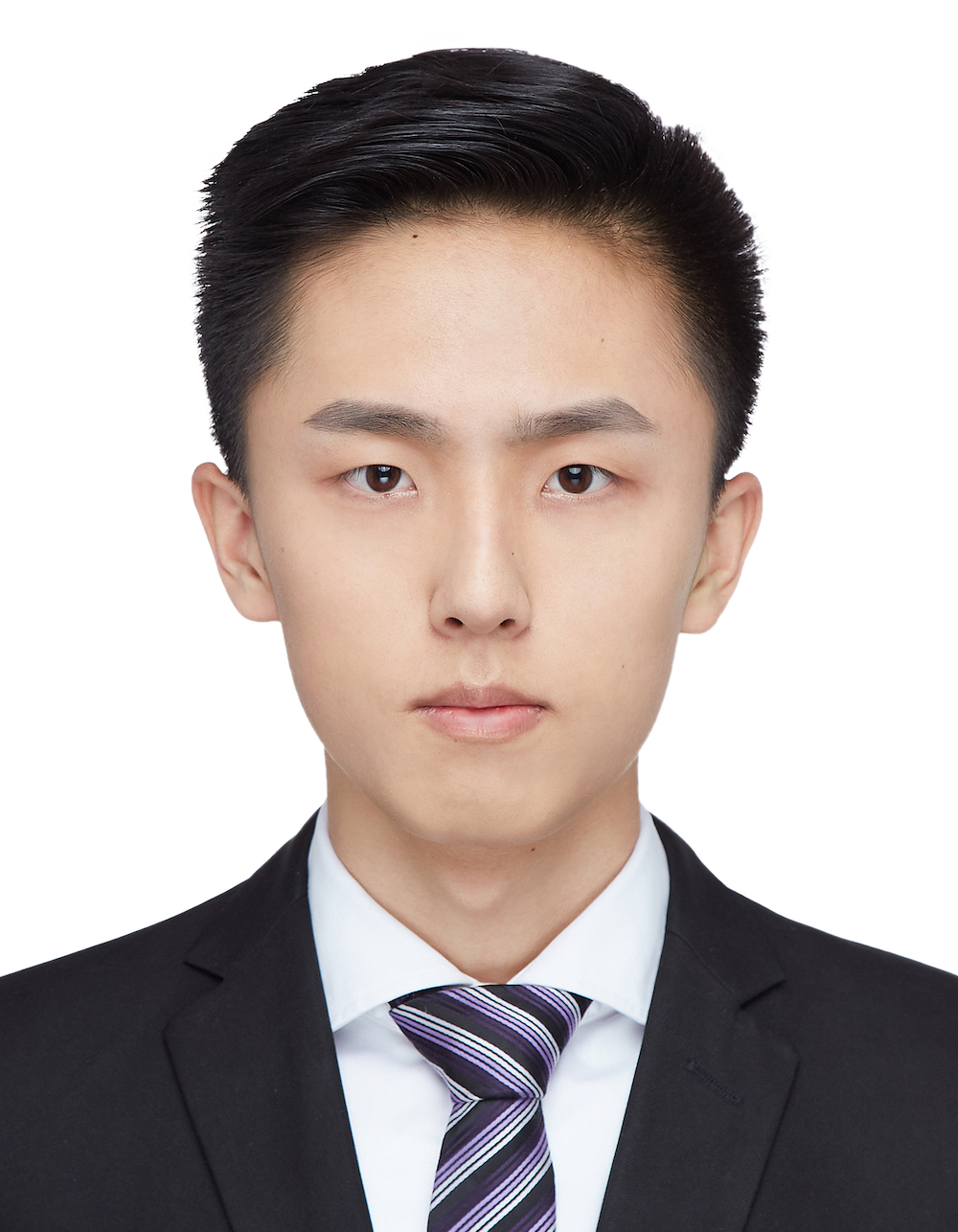}}]{Minghao Ning}
received the B.S. degree in vehicle engineering from the Beijing Institute of Technology, Beijing, China, in 2020, and completed his Ph.D. studies in mechanical and mechatronics engineering at the University of Waterloo, Waterloo, ON, Canada, in 2026. His research interests include autonomous driving, cooperative perception, socially aware robot navigation, and motion planning and control.
\end{IEEEbiography}

\begin{IEEEbiography}[{\includegraphics[width=1in,height=1.25in,clip,keepaspectratio]{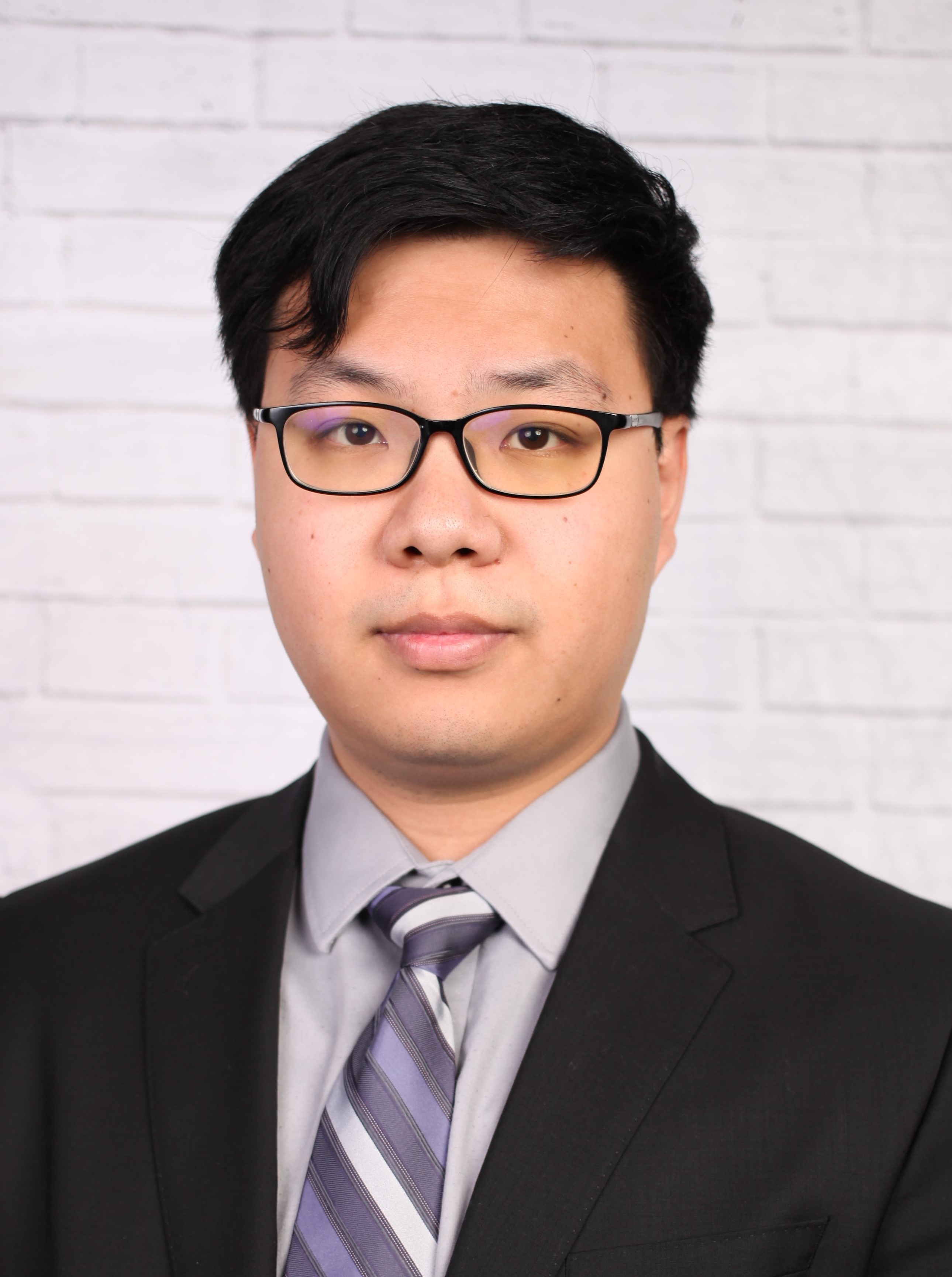}}]{Yufeng Yang}
is a Ph.D. candidate at the University of Waterloo Mechatronic Vehicle Systems (MVS) Lab. He received his B.Sc. degree in Mechanical Engineering with a minor in Mechatronics from the University of Calgary in 2021. His primary research interests include omnidirectional mobile robots, motion planning, control, and human-robot interaction.
\end{IEEEbiography}

\begin{IEEEbiography}[{\includegraphics[width=1in,height=1.25in,clip,keepaspectratio]{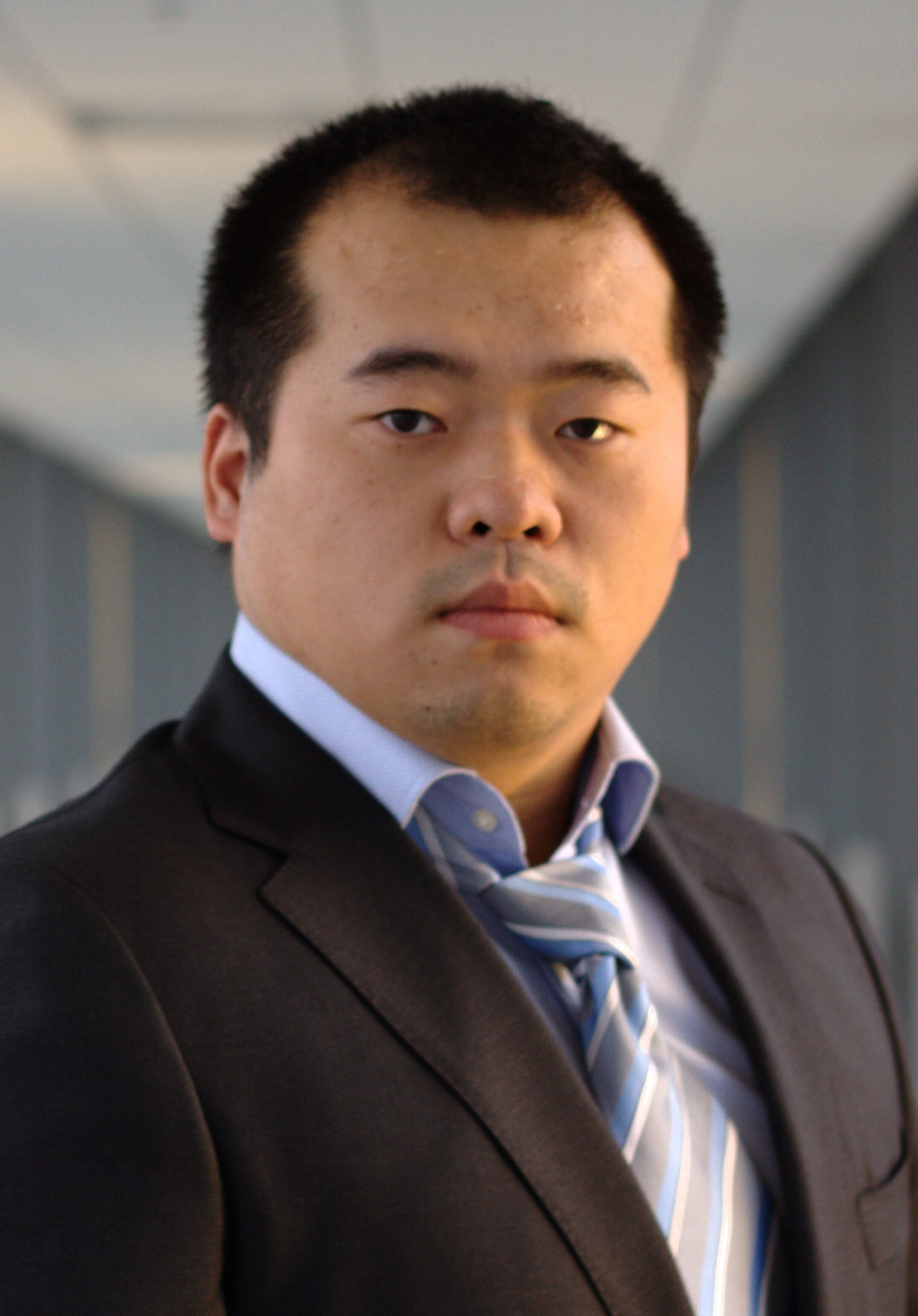}}]{Shucheng Huang} is currently a Ph.D. candidate at the University of Waterloo Mechatronic Vehicle Systems (MVS) Lab and CompLING Lab and a graduate student member at the Vector Institute.
He received the MASc degree in mechanical and mechatronics engineering
from the University of Waterloo in 2020 and the B.S. degree in mechanical engineering from Penn State University in 2018. His research interests include applications of LLM in autonomous driving, learning-based planning, and natural language processing.
\end{IEEEbiography}

\begin{IEEEbiography}
[{\includegraphics[width=1in,height=1.25in,clip,keepaspectratio]{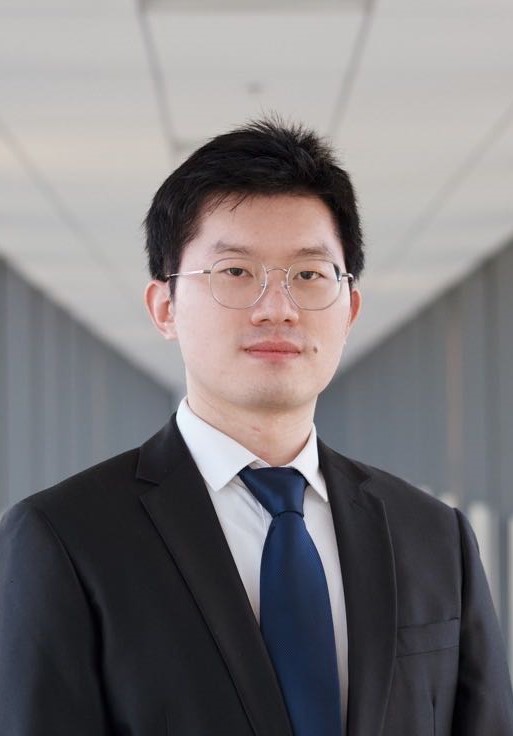}}]{Jiaming Zhong}
is currently a postdoc researcher at the University of Waterloo. He received his Ph.D. degree from the Mechatronic Vehicle Systems (MVS) Lab at the University of Waterloo in 2025. He was also a co-founder and the lead of planning and control of LoopX Innovation Inc. in Ontario, Canada. He received the B.S. and MASc degrees in mechanical engineering from Beijing Institute of Technology, China, in 2014 and 2017. His research interests include learning-based planning and control, multi-agent theory, and autonomous driving. 
\end{IEEEbiography}

\begin{IEEEbiography}[{\includegraphics[width=1in,height=1.25in,clip,keepaspectratio]{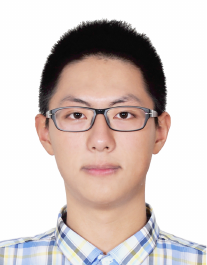}}]{Keqi Shu}
is a postdoctoral research fellow in Mechanical and Mechatronics Engineering at the University of Waterloo. His research interests involve interactive planning and decision-making of autonomous vehicles. 
He completed his PhD and Master's degree in Mechanical and Mechatronics Engineering in the University of Waterloo, Ontario, Canada, and his B.Sc. degree in Northwestern Polytechnical University, Xi'an Shaanxi, China.
\end{IEEEbiography}

\begin{IEEEbiography}[{\includegraphics[width=1in,height=1.25in,clip,keepaspectratio]{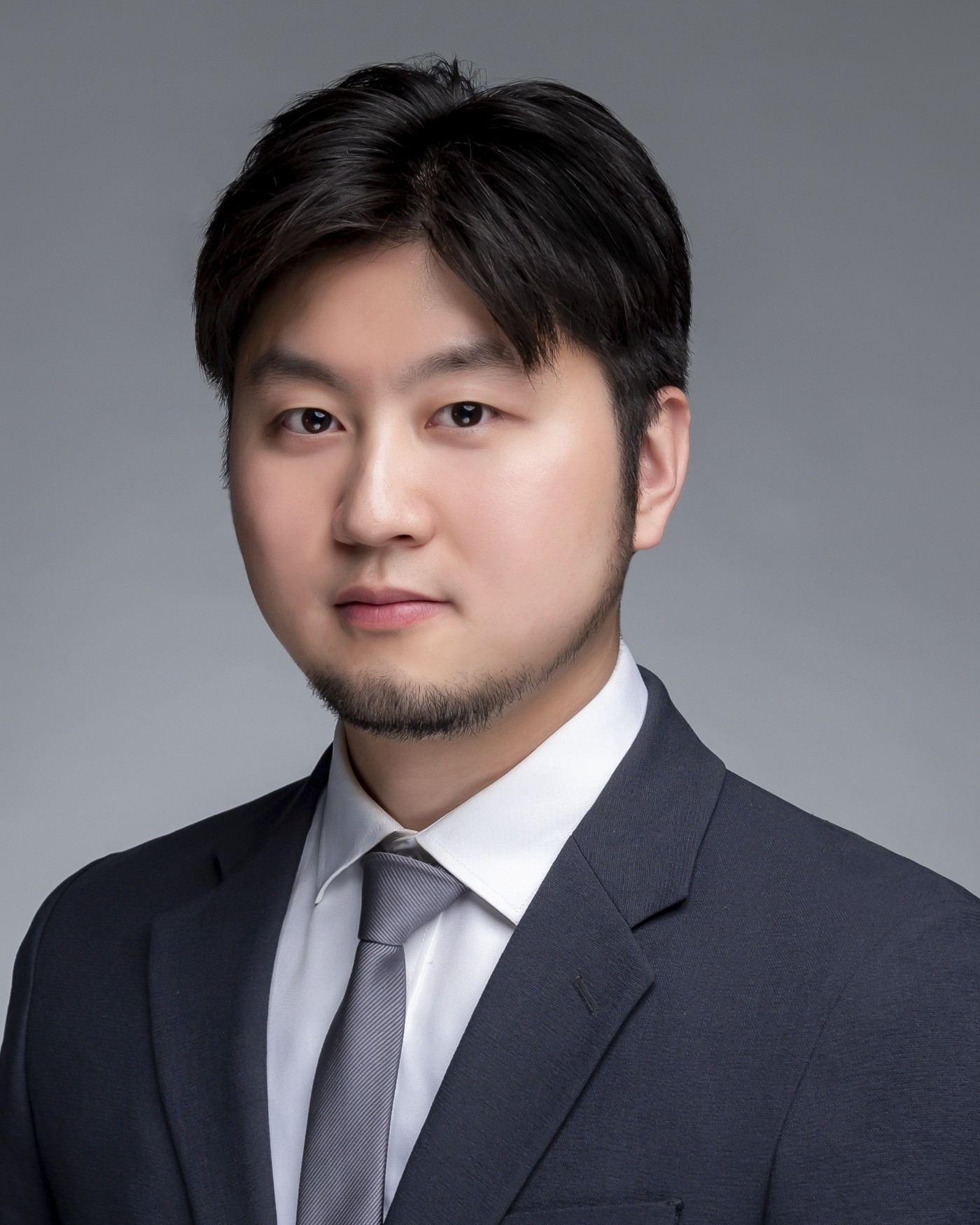}}]{Chen Sun}
received the Ph.D. degree in Mechanical
\& Mechatronics Engineering from University of
Waterloo, ON, Canada in 2022, M.A.Sc degree in
Electrical \& Computer Engineering from University
of Toronto, ON, Canada in 2017 and B.Eng. degree
in automation from the University of Electronic
Science and Technology of China, Chengdu, China,
in 2014. He is currently an Assistant Professor
with the Department of Data and Systems Engineering, University of Hong Kong. His research
interests include field robotics, safe and trustworthy
autonomous driving and in general human-CPS autonomy.
\end{IEEEbiography}

\begin{IEEEbiography}[{\includegraphics[width=1in,height=1.25in,clip,keepaspectratio]{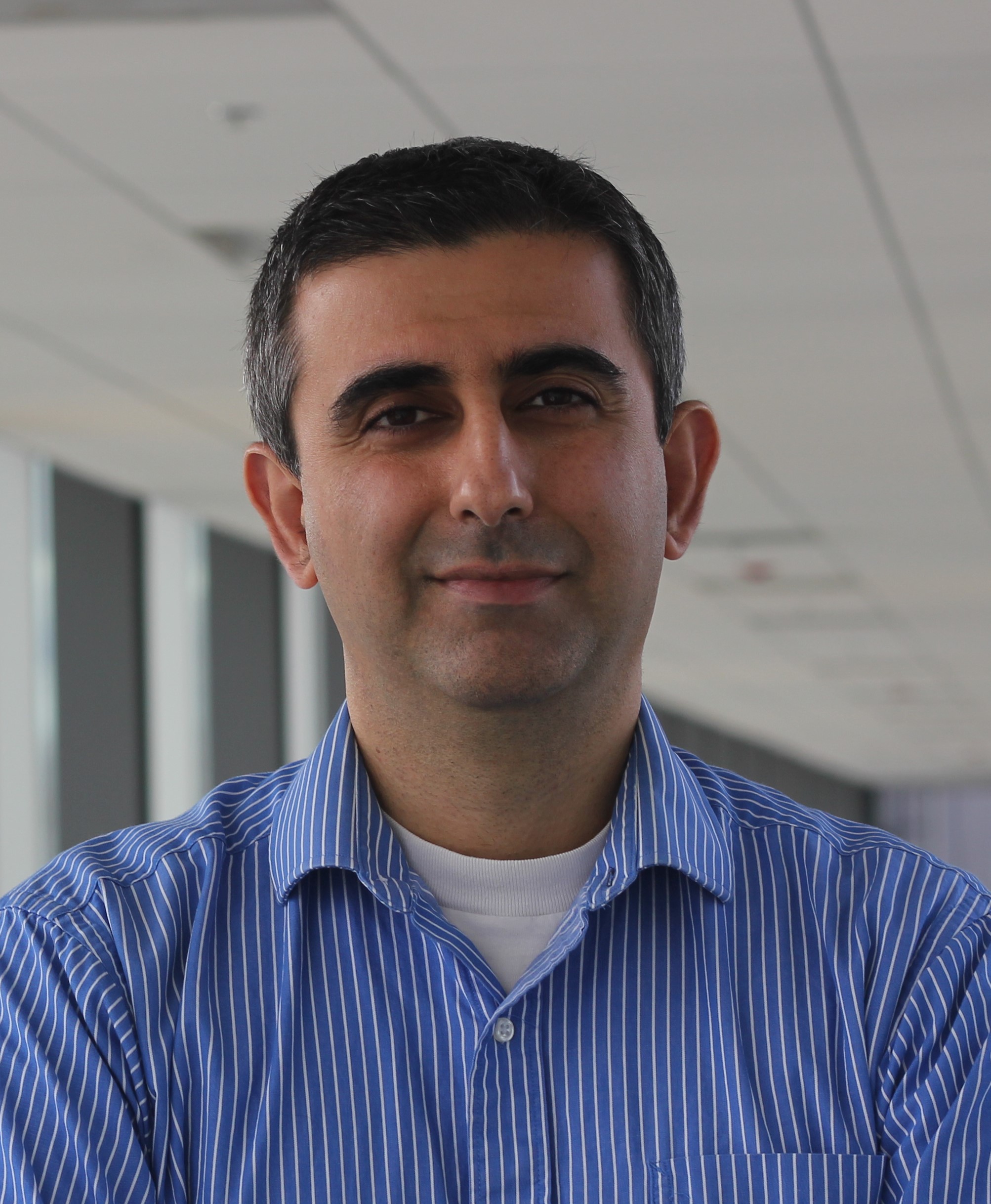}}]{Ehsan Hashemi}
received his Ph.D. in Mechanical and Mechatronics Engineering in 2017 from the University of Waterloo, ON, Canada; M.Sc. in Mechanical Engineering in 2005 from Amirkabir University of Technology (Tehran Polytechnic). He is currently an Assistant Professor at the Department of Mechanical Engineering, University of Alberta. His research interests are robotics, control theory, distributed estimation, and human-robot interaction.
\end{IEEEbiography}

\begin{IEEEbiography}[{\includegraphics[width=1in,height=1.25in,clip,keepaspectratio]{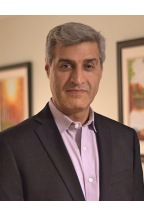}}]{Amir Khajepour}
is a professor of Mechanical and Mechatronics Engineering and the Director of the Mechatronic Vehicle Systems (MVS) Lab at the University of Waterloo. He held the Tier 1 Canada Research Chair in Mechatronic Vehicle Systems from 2008 to 2022 and the Senior NSERC/General Motors Industrial Research Chair in Holistic Vehicle Control from 2017 to 2022. His work has led to the training of over 150 PhD and MASc students, filing over 30 patents, publication of 600 research papers, numerous technology transfers, and the establishment of several start-up companies. He has been recognized with the Engineering Medal from Professional Engineering Ontario and is a fellow of the Engineering Institute of Canada, the American Society of Mechanical Engineering, and the Canadian Society of Mechanical Engineering.
\end{IEEEbiography}

\end{document}